\documentclass{article}

\PassOptionsToPackage{sort&compress,numbers}{natbib}

    \usepackage[final]{neurips_2021}

\usepackage[utf8]{inputenc} 
\usepackage[T1]{fontenc}    
\usepackage{hyperref}       
\usepackage{url}            
\usepackage{booktabs}       
\usepackage{amsfonts}       
\usepackage{nicefrac}       
\usepackage{microtype}      
\usepackage{xcolor}         
\usepackage{amsmath,graphicx,amssymb,amsthm}
\usepackage{adjustbox}
\usepackage{bm}
\usepackage{booktabs}
\usepackage{color}
\usepackage{multicol,multirow}
\usepackage{bbding}
\usepackage{cite}
\usepackage{hyperref}       
\usepackage{pdfpages}
\usepackage{comment}
\usepackage{framed}
\usepackage[most]{tcolorbox}
\usepackage{xcolor}
\colorlet{shadecolor}{orange!15}

\usepackage[ruled,vlined]{algorithm2e}

\DeclareMathOperator{\tr}{tr}

\DeclareMathOperator*{\argmin}{arg\,min}

\newcommand{\ali}[1]{\textcolor{black}{#1}}
\newcommand{\aliComment}[1]{\textcolor{red}{#1}}

\newtheorem{lem}{Lemma}
\newtheorem{prop}{Proposition}

\newtheorem{theo}[lem]{Theorem}
\newtheorem{cor}{Corollary}

\newtheorem{defi}{Definition}
\newtheorem{rem}{Remark}

\title{Efficient Hierarchical Bayesian Inference for Spatio-temporal Regression Models in Neuroimaging}

\vspace{-4mm}
\author{{\small Ali Hashemi$^{1,2}$, Yijing Gao$^{3}$, Chang Cai$^{3,4}$, Sanjay Ghosh$^{3}$,} \\ {\small \textbf{Klaus-Robert M{\"u}ller}$^{2,5,6,7}$, \textbf{Srikantan~S.~Nagarajan}$^{3}$, \textbf{and Stefan Haufe}$^{1,8,9,10}$} \\
{\small 
$^{1}$Uncertainty, Inverse Modeling and Machine Learning Group, Technische Universit{\"a}t Berlin, Germany.} \\ 
{\small $^{2}$Machine Learning Group, Technische Universit{\"a}t Berlin, Germany.}\\ 
{\small $^{3}$Department of Radiology and Biomedical Imaging, University of California, San Francisco, USA.} \\
{\small $^{4}$National Engineering Research Center for E-Learning, }{\small Central China Normal University, China.}\\
{\small $^{5}$BIFOLD -- Berlin Institute for the Foundations of Learning and Data, Berlin, Germany.} \\
{\small $^{6}$Department of Artificial Intelligence, Korea University, South Korea.} \\
{\small $^{7}$Max Planck Institute for Informatics, Saarbr{\"u}cken, Germany.} \\
{\small $^{8}$Physikalisch-Technische Bundesanstalt, Berlin, Germany.} \\
{\small $^{9}$Charit{\'e} -- Universit{\"a}tsmedizin Berlin, Germany.} \\
{\small $^{10}$Bernstein Center for Computational Neuroscience, Berlin, Germany.} \\
}

\begin{document}

\maketitle
\vspace{-3mm}
\begin{abstract}
Several problems in neuroimaging and beyond require inference on the parameters of multi-task sparse hierarchical regression models. Examples include M/EEG inverse problems, neural encoding models for task-based fMRI analyses, and climate science. In these domains, both the model parameters to be inferred and the measurement noise may exhibit a complex spatio-temporal structure. Existing work either neglects the {\em temporal} structure or leads to computationally demanding inference schemes. Overcoming these limitations, we devise a novel flexible hierarchical Bayesian framework within which the spatio-temporal dynamics of model parameters and noise are modeled to have Kronecker product covariance structure. Inference in our framework is based on majorization-minimization optimization and has guaranteed convergence properties. Our highly efficient algorithms exploit the intrinsic Riemannian geometry of temporal autocovariance matrices. For stationary dynamics described by Toeplitz matrices, the theory of circulant embeddings is employed. We prove convex bounding properties and derive update rules of the resulting algorithms. On both synthetic and real neural data from M/EEG, we demonstrate that our methods lead to improved performance.
\end{abstract}
\vspace{-2mm}
\section{Introduction}
\label{sec:intro}
%
Probabilistic graphical models for regression problems, where both the model parameters and the observation noise can have complex spatio-temporal correlation structure, are prevalent in neuroscience, neuroimaging and beyond \citep{shiffrin2008survey}. Example domains where these models are being applied include Gaussian process inference \citep{rakitsch2013all}, sparse signal recovery and multi-task learning \citep{candes2006stable,donoho2006compressed,bertrand2019handling,chevalier2020statistical}, array signal processing \citep{malioutov2005sparse}, fMRI analysis \citep{cai2020incorporating}, climate science \citep{beirle2003weekly}, computer hardware monitoring \citep{ranieri2015near}, and brain-computer interfaces (e.g.~\citep{lemm2005spatio,NIPS2014_fa83a11a}). The focus in this paper will be on brain source imaging (BSI) \citep{wu2016bayesian}, i.e. the reconstruction of brain source activity from non-invasive magneto- or electroetoencephalography (M/EEG) sensors \citep{baillet2001electromagnetic}. Although a number of generative models that encompass  spatio-temporal correlations have been proposed within the BSI regression problem, inference solutions have either imposed specific simplifications and model constraints \citep{cai2016bayesian,shvartsman2018matrix,cai2019representational} or have ignored temporal correlations overall \citep{bertrand2019handling}, and therefore not fully addressed the inherent spatio-temporal problem structure. 


We propose a novel and efficient algorithmic framework using Type-II Bayesian regression that explicitly considers the spatio-temporal covariance structure in both model coefficients and observation noise. To this end, we focus on the M/EEG inverse problem of brain source reconstruction, for which we adopt a multi-task regression approach and formulate the source reconstruction problem as a probabilistic generative model with Gaussian scale mixture priors for sources with spatio-temporal prior covariance matrices that are expressed as a Kronecker product of separable spatial and temporal covariances. Their solutions are constrained to the Riemannian manifold of positive definite (P.D.) temporal covariance matrices. Exploiting the concept of geodesic convexity on the Riemannian manifold of covariance matrices with Kronecker structure, we are then able to derive robust, fast and efficient \emph{majorization-minimization} (MM) optimization algorithms for model inference with provable convergence guarantees. In addition to deriving update rules for full-structural P.D. temporal covariance matrices, we also show that assuming a Toeplitz structure for the temporal covariance matrix results in computationally efficient update rules within the proposed MM framework.

\section{Spatio-temporal generative model}
\label{sec:gen-model}

Let us consider a multi-task linear regression problem, mathematically formulated as
\begin{align}
    {\bf Y}_{g} = {\bf L}{\bf X}_{g} + {\bf E}_{g} \quad \mathrm{for}\;g=1,\hdots,G \;,\label{eq:linearmodel}
\end{align}
in which a forward matrix, ${\bf L} \in \mathbb{R}^{M\times N}$, maps a set of coefficients or source components ${\bf X}_{g} \in\mathbb{R}^{N\times T}$ to measurements ${\bf Y}_{g} \in \mathbb{R}^{M\times T}$, with independent Gaussian noise ${\bf E}_{g} \in\mathbb{R}^{M\times T}$. \ali{$G$ denotes the number of sample blocks in the multi-task problem, while $N$, $M$, and $T$ denote the number of sources or coefficients, the number of sensors or observations, and the number of time instants, respectively.}
The problem of estimating $\{{\bf X}_{g}\}_{g=1}^{G}$ given ${\bf L}$ and $\{{\bf Y}_{g}\}_{g=1}^{G}$ can represent an inverse problem in physics, a multi-task regression problem in machine learning, or a multiple measurement vector (MMV) recovery problem in signal processing \citep{cotter2005sparse}. 


In the context of BSI, $\{{\bf Y}_{g}\}_{g=1}^{G}$ refers to M/EEG sensor measurement data, and $\{{\bf X}_{g}\}_{g=1}^{G}$ refers to brain source activity contributing to the observed sensor data. Here, $G$ can be defined as the number of epochs, trials or experimental tasks. The goal of BSI is to infer the underlying brain activity from the M/EEG measurements given the lead field matrix ${\bf L}$. In practice, ${\bf L}$ can be computed using discretization methods such as the finite element method for a given head geometry and known electrical conductivities using the quasi-static approximation of Maxwell's equations
\citep{hamalainen1993magnetoencephalography}. 
As the number of locations of potential brain sources is dramatically larger than the number of sensors, the M/EEG inverse problem is highly ill-posed, which can be dealt with by incorporating prior assumptions. Adopting a Bayesian treatment is useful in this capacity because it allows these assumptions to be made explicitly by specifying prior distributions for the model parameters. Inference can be performed either through Maximum-a-Posteriori (MAP) estimation (\emph{Type-I Bayesian learning}) \citep{pascual1994low,haufe2008combining,gramfort2012mixed} or, when the model has unknown hyperparameters, through Type-II Maximum-Likelihood (ML) estimation (\emph{Type-II Bayesian learning}) \citep{mika2000mathematical,tipping2001sparse,wipf2009unified}. 

Here we focus on full spatio-temporal Type-II Bayesian learning, which assumes a family of prior distributions $p({\bf X}_{g}|{\bm \Theta})$ parameterized by a set of hyperparameters $\bm \Theta$. We further assume that the spatio-temporal correlation structure of the brain sources can be modeled by a Gaussian $p({\bf x}_{g}|{\bm \Gamma},{\bf B})\sim\mathcal{N}({\bm 0},{\bm \Sigma}_{0})$, where ${\bf x}_{g}=\text{vec}({\bf X}_{g}^{\top}) \in \mathbb{R}^{NT \times 1}$ and where the covariance matrix ${\bm \Sigma}_{0}={\bm \Gamma} \otimes {\bf B}$ has Kronecker structure, meaning that the temporal and spatial correlations of the sources are modeled independently through  matrices ${\bm \Gamma}$ and ${\bf B}$, where $\otimes$ stands for the Kronecker product. Note 
that the decoupling of spatial and temporal covariance is a neurophysiologically plausible assumption as neuronal sources originating from different parts of the brain can be assumed to share a similar autocorrelation spectrum. In this paper, we set ${\bm \Gamma}=\mathrm{diag}({\bm \gamma})$, ${\bm \gamma} = [\gamma_{1},\dots,\gamma_{N}]^\top$, and ${\bf B} \in \mathbb{S}^{T}_{++}$, where $\mathbb{S}^{T}_{++}$ denotes the set of positive definite matrices with size $T \times T$. Note that using a diagonal matrix ${\bm \Gamma}$ amounts to assuming that sources in different locations are independent. This assumption may be relaxed in our future work. On the other hand, modeling a full temporal covariance matrix ${\bf B}$ amounts to assuming non-stationary dynamics of the sources, which is appropriate in event-related experimental designs. To deal with the more stationary sources (ongoing brain activity), we will also constrain ${\bf B}$ to have Toeplitz structure later on. Based on these specifications, the prior distribution of the $i$-th brain source is modeled as $p(({\bf X}_{g})_{i.}|\gamma_{i},{\bf B})\sim\mathcal{N}(0,\gamma_{i}{\bf B}),\text{for}\; i=1,\dots,N$, where $({\bf X}_{g})_{i.}$ denotes the $i$-th row of source matrix ${\bf X}_g$. 

Analogous to the definition of the sources, we here also model the noise to have spatio-temporal structure. The matrix ${\bf E}_{g} \in \mathbb{R}^{M\times T}$ represents $T$ time instances of zero-mean Gaussian noise with full covariance, ${\bm \Sigma}_{\mathrm{\bf e}}={\bm \Lambda} \otimes {\bm \Upsilon}$, where ${\bf e}_{g}=\text{vec}({\bf E}_{g}^{\top}) \in \mathbb{R}^{MT\times1}\sim \mathcal{N}(0,{\bm \Sigma}_{\mathrm{\bf e}})$, and where ${\bm \Lambda}$ and ${\bm \Upsilon}$ denote spatial and temporal noise covariance matrices, respectively. Here, we assume that the noise and sources share the same temporal structure, i.e., ${\bm \Upsilon}={\bf B}$. We later investigate violations of this assumption in the simulation section. Analogous to the sources, we consider spatially independent noise characterized by diagonal spatial covariance ${\bm \Lambda} =\mathrm{diag}({\bm \lambda})$, ${\bm \lambda} = [\lambda_{1},\dots,\lambda_{M}]^\top$. We consider both the general, \emph{heteroscedastic}, case in which the noise level may be different for each sensor as well as the \emph{homoscedastic} case in which the noise level is uniform. 

For later use, we also define augmented versions of the source and noise covariance matrices as well as the lead field. Specifically, we set ${\bf H}$ and ${\bm \Phi}$ so that ${\bf H}:= [{\bm \Gamma}, {\bf 0}; {\bf 0}, {\bm \Lambda}]$, and ${\bm \Phi}:=[{\bf L},{\bf I}]$. These definitions allow us to unify source and noise covariance parameters within the single variable ${\bf H}$, which facilitates the concurrent estimation of both quantities. ${\bm \Phi}$ now plays the same role as the lead field ${\bf L}$, in the sense that it maps ${\bm \eta}_g:=[{\bf x}_g^{\top},{\bf e}_g^{\top}]$ to the measurements; i.e., ${\bf y}_g = {\bm \Phi}{\bm \eta}_g$. \ali{Figure~\ref{fig:graphprob-geometricmean}-(a) illustrates a probabilistic graphical model summarizing the spatio-temporal generative model of our multi-task linear regression problem.}

The MMV model Eq.~\eqref{eq:linearmodel} can be formulated equivalently in \emph{single measurement vector (SMV)} form by vectorizing the spatio-temporal data matrices and using the Kronecker product as follows: ${\bf y}_{g}={\bf D}{\bf x}_{g}+{\bf e}_{g}$, where ${\bf y}_{g}=\text{vec}\left({\bf Y}_{g}^\top \right)\in\mathbb{R}^{MT\times1}$, and ${\bf D}={\bf L}\otimes {\bf I}_{T}$. In a Type-II Bayesian learning framework, the hyper-parameters of the spatio-temporal source model are optimized jointly with the model parameters $\{\bm X_g \}_{g=1}^G$. In our case, these hyper-parameters comprise the unknown source, noise and temporal covariance matrices, i.e., ${\bm \Theta} = \{{\bm \Gamma},{\bm \Lambda},{\bf B}\} $. 
\section{Proposed method --- full Dugh}
\label{sec:full-dugh}
The unknown parameters ${\bm \Gamma}$, ${\bm \Lambda}$, and ${\bf B}$ can be optimized in an alternating iterative process. Given initial estimates, the posterior distribution of the sources is a Gaussian of the form $p({\bf x}_{g}|{\bf y}_{g},{\bm \Gamma},{\bm \Lambda},{\bf B}) \sim \mathcal{N}(\bar{\bf x}_{g},{\bm \Sigma}_{\bf x})$, whose mean and covariance are defined as follows:
\begin{align}
    \bar{\bf x}_{g} &={\bm \Sigma}_{0}{\bf D}^{\top}({\bm \Lambda} \otimes {\bf B} + {\bf D}{\bm \Sigma}_{0}{\bf D}^{\top})^{-1}{\bf y}_{g}={\bm \Sigma}_{0}{\bf D}^{\top}\Tilde{\bm \Sigma}_{\bf y}^{-1}{\bf y}_{g}\;,
    \label{eq:MeanValue-ST} \\
    {\bm \Sigma}_{\bf x} &= {\bm \Sigma}_{0}-{\bm \Sigma}_{0}{\bf D}^{\top}\Tilde{\bm \Sigma}_{\bf y}^{-1}{\bf D}{\bm \Sigma}_{0}
    \label{eq:CovarianceValue-ST}\;,
\end{align}
where ${\bm \Sigma}_{\bf y}={\bf L}{\bm \Gamma}{\bf L}^{\top}+{\bm \Lambda}$, and where $\Tilde{\bm \Sigma}_{\bf y} = {\bm \Sigma}_{\bf y} \otimes {\bf B}$ denotes the spatio-temporal statistical model covariance matrix. The estimated posterior parameters $\bar{\bf x}_{g}$ and ${\bm \Sigma}_{\bf x}$ are then in turn used to update ${\bm \Gamma}$, ${\bm \Lambda}$, and $\bf B$ as the minimizers of the negative log of the marginal likelihood $p({\bf Y}|{\bm \Gamma},{\bm \Lambda},{\bf B})$, which is given by:
\begin{align}
    \mathcal{L}_{\text{kron}}({\bm \Gamma},{\bm \Lambda},{\bf B}) &= T\log|{{\bm \Sigma}_{\bf y}}|+M\log|{\bf B}| +\frac{1}{G}\sum_{g=1}^{G}\text{tr}({{\bm \Sigma}_{\bf y}^{-1}}{\bf Y}_{g}{\bf B}^{-1}{\bf Y}_{g}^{\top}) 
    \label{eq:MLKronokCost}\;,
\end{align}
where $|\cdot|$ denotes the determinant of a matrix. A detailed derivation is provided in Appendix~\ref{appendix:spatio-temp-cost}. Given the solution of the hyperparameters ${\bm \Gamma}$, ${\bm \Lambda}$, and $\bf B$, the posterior source distribution is obtained by plugging these estimates into Eqs.~\eqref{eq:MeanValue-ST}--\eqref{eq:CovarianceValue-ST}. This process is repeated until convergence.

The challenge in this high-dimensional inference problem is to find (locally) optimal solutions to  Eq.~\eqref{eq:MLKronokCost}, which is a non-convex cost function, in adequate time. Here we propose a novel efficient algorithm, which is able to do so. Our algorithm thus learns the full spatio-temporal correlation structure of sources and noise. \citet{hashemi2021unification} have previously demonstrated that 
\emph{majorization-minimization} (MM) \citep{sun2017majorization}
is a powerful non-linear optimization framework that can be leveraged to solve similar Bayesian Type-II inference problems. Here we extend this work to our spatio-temporal setting. Building on the idea of majorization-minimization, we construct convex surrogate functions that \emph{majorize} $\mathcal{L}_{\text{kron}}({\bm \Gamma},{\bm \Lambda},{\bf B})$ in each iteration of the proposed algorithm. Then, we show the minimization equivalence between the constructed majorizing functions and Eq.~\eqref{eq:MLKronokCost}. These results are presented in theorems~\ref{Theo:Time-Surrogate} and \ref{Theo:Space-Surrogate}. Theorems~\ref{Theo:Time-Sol} and \ref{Theo:Space-Sol} propose an efficient alternating optimization algorithm for solving $\mathcal{L}_{\text{kron}}({\bm \Gamma},{\bm \Lambda},{\bf B})$, which leads to update rules for the spatial and temporal covariance matrices ${\bm \Gamma}$ and ${\bf B}$ as well as the noise covariance matrix ${\bm \Lambda}$. 

Starting with the estimation of the temporal covariance based on the current source estimate, we can state the following theorem: 
%
\begin{figure}
\begin{center}
\begin{minipage}[b]{0.49\textwidth}
    \centerline{\includegraphics[width=7.0cm,keepaspectratio]{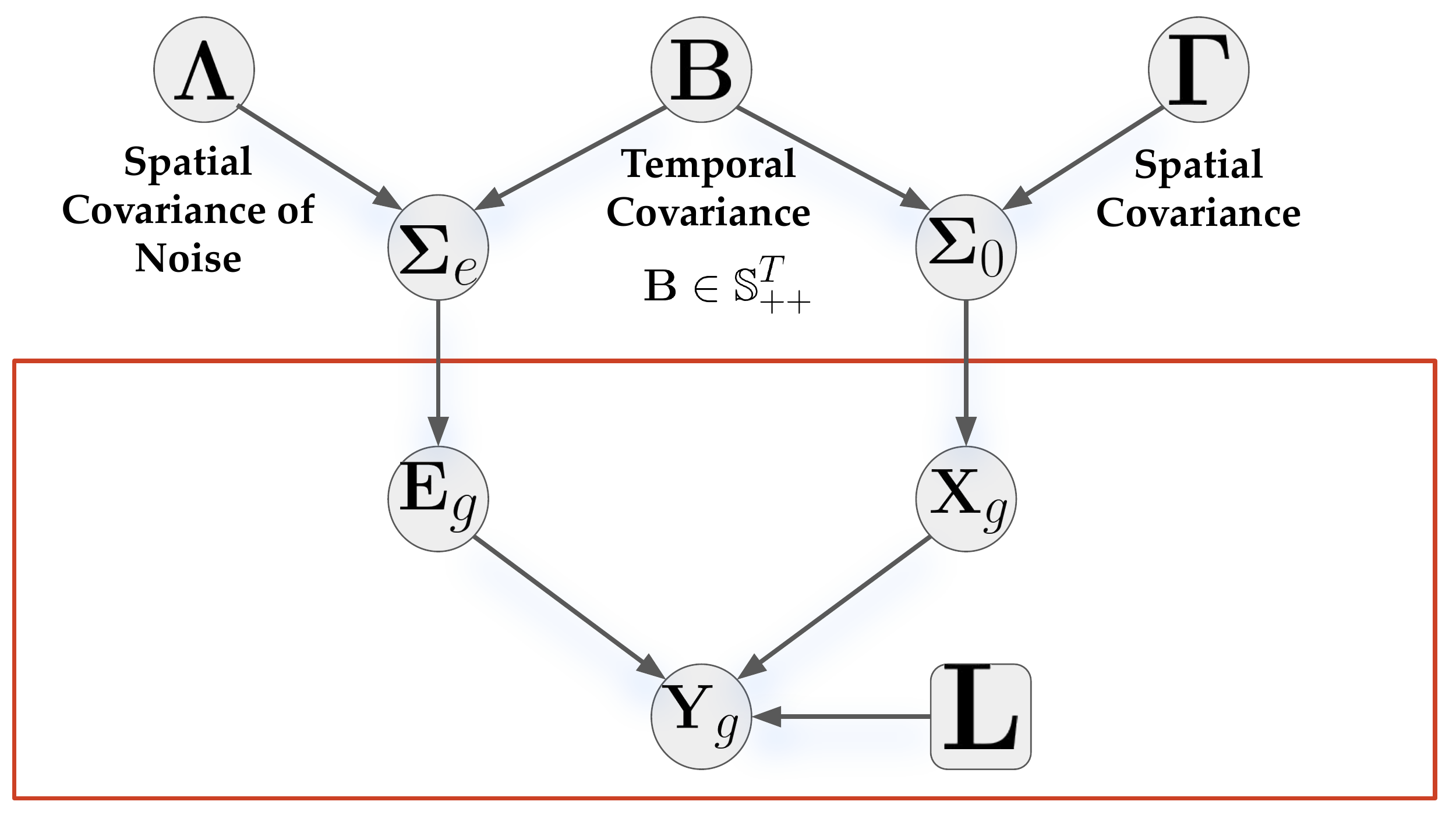} } 
    \begin{center}
        \centerline{(a)}
    \end{center}
\end{minipage}
\begin{minipage}[b]{0.49\textwidth}
    \centerline{\includegraphics[width=7.0cm,keepaspectratio]{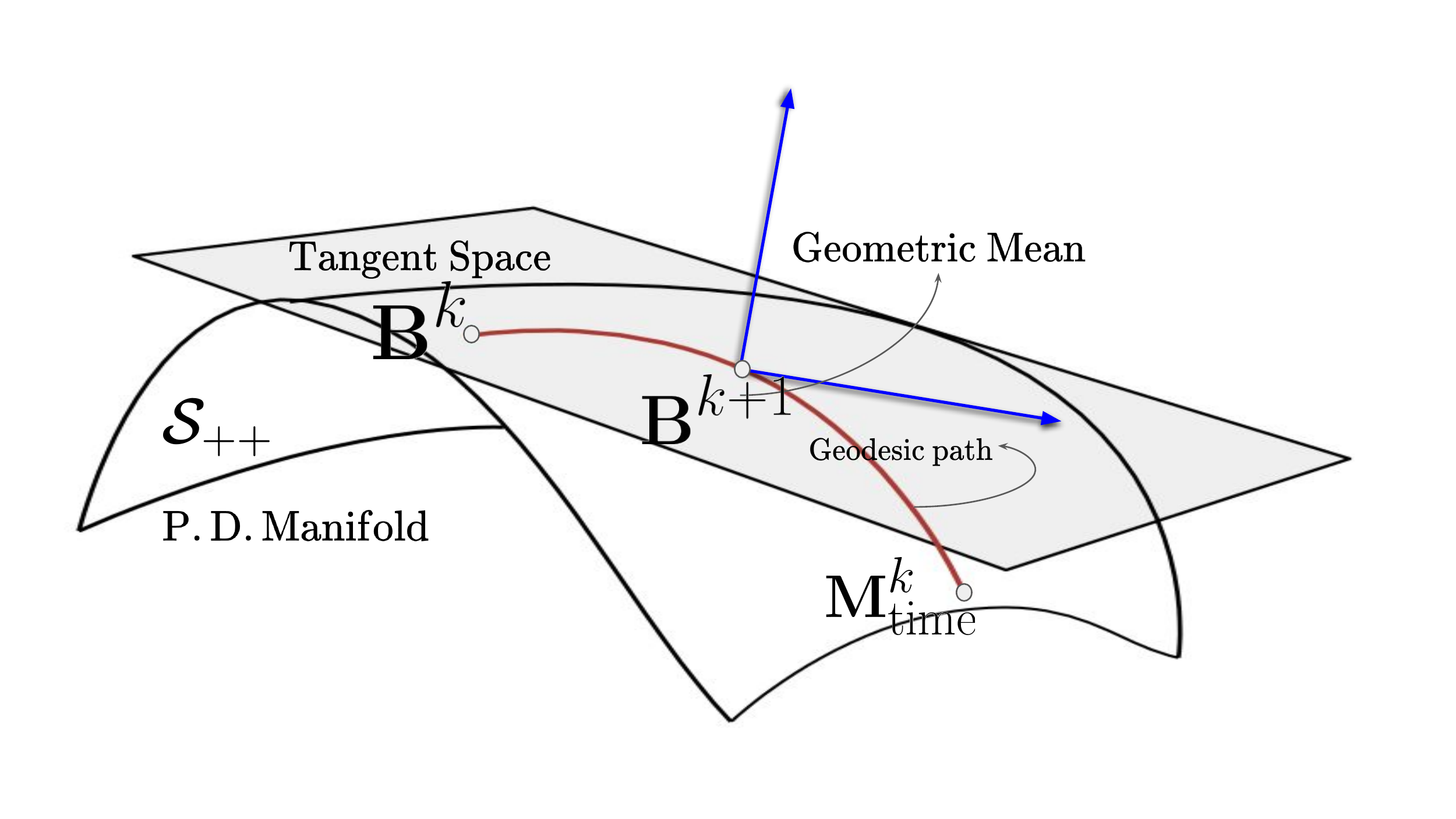}} 
    \begin{center}
        \centerline{(b)}
    \end{center}
\end{minipage}
\end{center}
\caption{\ali{(a) Probabilistic graphical model for the multi-task linear regression problem. (b) Geometric representation of the geodesic path between the pair of matrices 
$\{{\bf B}^{k},{\bf M}_{\mathrm{time}}^{k}\}$ on the P.D. manifold and the geometric mean between them, which is used to update ${\bf B}^{k+1}$.}}
\label{fig:graphprob-geometricmean}
\vspace{-3mm}
\end{figure}

\begin{theo} 
\label{Theo:Time-Surrogate}
Optimizing the non-convex Type-II cost function in Eq.~\eqref{eq:MLKronokCost},  $\mathcal{L}_{\text{kron}}({\bm \Gamma},{\bm \Lambda},{\bf B})$, with respect to the temporal covariance matrix ${\bf B}$ is equivalent to optimizing the following convex surrogate function, which \emph{majorizes} Eq.~\eqref{eq:MLKronokCost}:  
\begin{align}
    \mathcal{L}^{\mathrm{time}}_{\mathrm{conv}}({\bm \Gamma}^{k},{\bm \Lambda}^{k},{\bf B})=\tr\left(({\bf B}^{k})^{-1}{\bf B}\right)+\tr({\bf M}_{\mathrm{time}}^{k}{\bf B}^{-1}),
    \label{eq:TimeSurrogateFunction}    
\end{align}
where ${\bf M}_{\mathrm{time}}^{k}$ is defined as
\begin{equation}
    {\bf M}_{\mathrm{time}}^{k}:=\frac{1}{MG}\sum_{g=1}^{G}{\bf Y}_{g}^{\top}\left({\bm \Sigma}_{\bf y}^{k}\right)^{-1}{\bf Y}_{g}
    \label{eq:TempSampleCov}\;,
\end{equation}
and where ${\bm \Gamma}^{k}$, ${\bm \Lambda}^{k}$, and ${\bm \Sigma}_{\bf y}^{k}$ denote the source, noise and statistical model covariance matrices at the $k$-th iteration, which are treated as constants when optimizing over ${\bf B}$. 
\end{theo}
\begin{proof}
A detailed proof is provided in Appendix \ref{appendix:Time-Surrogate}.
\end{proof}
The solution of Eq.~\eqref{eq:TimeSurrogateFunction} can be obtained by taking advantage of the Riemannian geometry of the search space. The following theorem holds. 
\begin{theo}
\label{Theo:Time-Sol}
The cost function $\mathcal{L}^{\mathrm{time}}_{\mathrm{conv}}({\bm \Gamma}^{k},{\bm \Lambda}^{k},{\bf B})$ is strictly geodesically convex with respect to the P.D. manifold and its minimum with respect to ${\bf B}$ can be attained by iterating the following update rule until convergence:
\begin{align}
    {\bf B}^{k+1} \leftarrow ({\bf B}^{k})^{\nicefrac{1}{2}}\left(({\bf B}^{k})^{\nicefrac{-1}{2}}{\bf M}_{\mathrm{time}}^{k}({\bf B}^{k})^{\nicefrac{-1}{2}}\right)^{\nicefrac{1}{2}}({\bf B}^{k})^{\nicefrac{1}{2}}\;\label{eq:BGeodesicUpdate}.
\end{align}
\end{theo}
\begin{proof}
\ali{A detailed proof is provided in Appendix \ref{appendix:Time-Sol}.}
\end{proof}

%
\begin{rem}
    A geometric interpretation of the update rule \eqref{eq:BGeodesicUpdate} is that it finds the geometric mean between a spatially whitened version of the empirical temporal covariance matrix, ${\bf M}_{\mathrm{time}}^{k}$, and the model temporal covariance matrix from the previous iteration, ${\bf B}^{k}$. A geometric representation of the geodesic path between the pair of matrices $\{{\bf B}^{k},{\bf M}_{\mathrm{time}}^{k}\}$ on the P.D. manifold and the geometric mean between them, representing the update for ${\bf B}^{k+1}$, is provided in Figure~\ref{fig:graphprob-geometricmean}-(b).
\end{rem}
We now derive an update rule for the spatial source and noise covariance matrices using an analogous approach. To this end, we first construct a convex surrogate function that \emph{majorizes} $\mathcal{L}_{\text{kron}}({\bm \Gamma},{\bm \Lambda},{\bf B})$ in each iteration of the optimization algorithm by considering augmented variables comprising both the source and noise covariances, i.e., ${\bf H}:= [{\bm \Gamma}, \bf 0; \bf 0, {\bm \Lambda}]$, and ${\bm \Phi}:=[{\bf L},{\bf I}]$. The following theorem holds. 
\begin{theo} 
\label{Theo:Space-Surrogate}
Minimizing the non-convex Type-II cost function in Eq.~\eqref{eq:MLKronokCost}, $\mathcal{L}_{\text{kron}}({\bm \Gamma},{\bm \Lambda},{\bf B})$, with respect to the spatial covariance matrix ${\bf H}$ is equivalent to minimizing the following convex surrogate function, which \emph{majorizes} Eq.~\eqref{eq:MLKronokCost}:
\begin{align}
    \mathcal{L}^{\mathrm{space}}_{\mathrm{conv}}({\bm \Gamma},{\bm \Lambda},{\bf B}^{k}) = \mathcal{L}^{\mathrm{space}}_{\mathrm{conv}}({\bf H},{\bf B}^{k})=\tr\left({\bm \Phi}^{\top}({\bm \Sigma}_{\bf y}^{k})^{-1}{\bm \Phi}{\bf H}\right)+\tr\left({\bf M}_{\mathrm{SN}}^{k} {\bf H}^{-1} \right) \;,
    \label{eq:SpaceSurrogateFunction}    
\end{align}
where ${\bf M}_{\mathrm{SN}}^{k}$ is defined as
\begin{align}
        {\bf M}_{\mathrm{SN}}^{k} &:= {\bf H}^{k}{\bm \Phi}^{\top}({\bm \Sigma}_{\bf y}^{k})^{-1}{\bf M}_{\mathrm{space}}^{k}({\bm \Sigma}_{\bf y}^{k})^{-1}{\bm \Phi} {\bf H}^{k} \;, \mathrm{with} \nonumber \\ 
        {\bf M}_{\mathrm{space}}^{k} &:= \frac{1}{TG}\sum_{g=1}^{G}{\bf Y}_{g}({\bf B}^{k})^{-1}{\bf Y}_{g}^{\top} \;,
        \label{eq:SpaceSampleCov}
\end{align}
and where ${\bf B}^k$ denotes the temporal model covariance matrix estimated in the $k$-th iteration.
\end{theo}
\begin{proof}
A detailed proof is provided in Appendix \ref{appendix:Space-Surrogate}.
\end{proof}


%
%
While, in principle, update rules for full-structured spatial source and noise covariances may be conceived, in analogy to the ones presented for the temporal covariance, we here restrict ourselves to the discussion of diagonal spatial covariances. Note, though, that such a choice of prior does not prohibit the reconstruction of parameters with more complex correlation structure. We have ${\bf H}=\mathrm{diag}({\bf h})$, ${\bf h} = [\gamma_{1},\dots,\gamma_{N}, \sigma^2_{1},\dots,\sigma^2_{M}]^\top$, and ${\bm \Sigma}_{\bf y}={\bm \Phi}{\bf H}{\bm \Phi}^{\top}$. The update rule for ${\bf H}$ then takes a simple form, as stated in the following theorem:
\begin{theo}
\label{Theo:Space-Sol}
The cost function $\mathcal{L}^{\mathrm{space}}_{\mathrm{conv}}({\bf H},{\bf B}^{k})$ is convex in $\bf h$, and its minimum with respect to $\bf h$ can be obtained according to the following closed-form update rule, which concurrently estimates the scalar source variances and heteroscedastic noise variances:
\begin{align}
    {\bf H}^{k+1}=\mathrm{diag}({\bf h}^{k+1}),\;   h_{i}^{k+1}&\leftarrow\sqrt{\frac{g_{i}^{k}}{z_{i}^{k}}}\quad \text{for}\;i=1,\hdots,N+M \;\label{eq:HDiagonalUpdate}, \text{where} \\ 
    {\bf g} &:=\mathrm{diag}({\bf M}_{\mathrm{SN}}^{k}) \label{eq:diagISpace}  \\
    {\bf z} &:=\mathrm{diag}({\bm \Phi}^{\top}({\bm \Sigma}_{\bf y}^{k})^{-1}{\bm \Phi}) \label{eq:diagIISpace} 
\end{align} 
\end{theo}
\begin{proof}
A detailed proof can be found in Appendix \ref{appendix:Space-Solution}.
\end{proof}
The final estimate of the posterior distribution \eqref{eq:MeanValue-ST} can be obtained by starting from any initial guess for the model parameters, ${\bf H}^{0}=\{{\bm \Gamma}^{0},{\bm \Lambda}^{0}\}$ and ${\bf B}^{0}$, and iterating between update rules \eqref{eq:BGeodesicUpdate}, \eqref{eq:HDiagonalUpdate}, and \eqref{eq:MeanValue-ST} until convergence. We call the resulting algorithm \emph{full Dugh}. Convergence is shown in the following theorem:
\begin{theo}
\label{theo:MM-Convergence-Guarantees}
Optimizing the non-convex ML cost function  Eq.~\eqref{eq:MLKronokCost} with alternating update rules for ${\bf B}$ and ${\bf H}=\{{\bm \Gamma},{\bm \Lambda}\}$ in Eq.~\eqref{eq:BGeodesicUpdate} and Eq.~\eqref{eq:HDiagonalUpdate} defines an MM algorithm, which is guaranteed to converge to a stationary point. 
\end{theo}
\begin{proof}
A detailed proof can be found in Appendix~\ref{appendix:MM-Convergence-Guarantees}.
\end{proof}
%
\section{Efficient inference for stationary temporal dynamics --- thin Dugh}
\label{sec:thin-Dugh}
With full Dugh, we could present a general inference method for data with full temporal covariance structure. This algorithm may be used for non-stationary data, for which the covariance between two time samples depends on the absolute position of the two samples within the studied time window. An example is the reconstruction of brain responses to repeated external stimuli from M/EEG (event-related electrical potential or magnetic field) data, where data blocks are aligned to the stimulus onset. In other realistic settings, however, stationarity can be assumed in the sense that the covariance between two samples only depends on their relative position (distance) to each other but not to any external trigger. An example is the reconstruction of ongoing brain dynamics in absence of known external triggers. This situation can be adequately modeled using temporal covariance matrices with Toeplitz structure. In the following, we devise an efficient extension of full Dugh for that setting. 

Temporal correlation has been incorporated in various brain source imaging models through different priors \citep[see][and references therein]{pirondini2017computationally,wu2016bayesian}. According to  \citep{lamus2012spatiotemporal}, the temporal dynamics of neuronal populations can be approximated by a first order auto-regressive (AR(1)) model of the form ${x}_n(t+1)=\beta {x}_n(t)+\sqrt{1-\beta^{2}}\xi_n(t),~ n=1,\hdots,N; t=1,\hdots,T$ with AR coefficient $\beta \in(-1,1)$ and innovation noise $\xi_n(t)$. It can be shown that the temporal correlation matrix corresponding to this AR model has Toeplitz structure, ${\bf B}_{i,j}=\beta^{|{i-j}|}$ \citep{zhang2011sparse}. Consequently,  we now constrain the cost function Eq.~\eqref{eq:TimeSurrogateFunction} to the set of Toeplitz matrices:
\begin{align}
    {\bf B}^{k+1} &=\argmin_{{\bf B} \in \mathcal{B},\;{\bf H}= {\bf H}^{k} } \tr(({\bf B}^{k})^{-1}{\bf B})+\tr({\bf M}_{\mathrm{time}}^{k}{\bf B}^{-1})\;,
    \label{eq:MinComstrainedTempo}
\end{align}
where set $\mathcal{B}$ denotes the set of real-valued positive-definite Toeplitz matrices of size $T\times T$. 

In order to be able to derive efficient update rules for this setting, we bring ${\bf B}$ into a diagonal form. The following proposition holds.
\begin{prop} 
\label{prop:toeplitz}
Let ${\bf P} \in \mathbb{R}^{L \times L}$ with $L>T$, be the circulant embedding of matrix ${\bf B} \in \mathcal{B}^L$, where $\mathcal{B}^L$ is assumed to be the subset of $\mathcal{B}$ that guarantees that ${\bf P}$ is a real-valued circulant P.D. matrix.  
Then the Toeplitz matrix ${\bf B}$ can be diagonalized as follows:
\begin{align}
    {\bf B} &= {\bf Q}{\bf P}{\bf Q}^{H} \quad \text{with} \quad \mathbf{Q} = 
    [{\bf I}_{M},{\bf 0}]{\bf F}_{L}   \label{eq:FourierDiagonalization},\;\text{and}  \\
    [\mathbf{F}_{L}]_{m,l} &= \frac{1}{\sqrt{L}}e^{\frac{i2\pi(l-1)}{L}(m-1)} \label{eq:DFT}\;,
\end{align}
where ${\bf P}=\mathrm{diag}({\bf p}) =\mathrm{diag}(p_{0},p_{1},\dots,p_{L-1})$ is a diagonal matrix. The main diagonal coefficients of ${\bf P}$ are given as the normalized discrete Fourier transform (DFT), represented by the linear operator $\mathbf{F}_{L}$, of the first row of ${\bf B}$: ${\bf p}=\mathbf{F}_{L}{\bf B}_{1.}$  
(Note that a Toeplitz matrix can be represented by its first row or column, ${\bf B}_{1.}$). 
\end{prop}
\begin{proof}
This is a direct implication of the fact that Toeplitz matrices can be embedded into circulant matrices of larger size \citep{dembo1989embedding} and the result that circulant matrices can be approximately diagonalized using the Fourier transform \citep{grenander1958toeplitz}. Further details can be found in Appendix~\ref{appendix:toeplitz}.
\end{proof}
The solution of Eq.~\eqref{eq:MinComstrainedTempo} can be obtained by direct application of Proposition~\ref{prop:toeplitz}. The results is summarized in the following theorem: 
\begin{theo}
\label{Theo:Time-Toeplitz}
The cost function Eq.~\eqref{eq:MinComstrainedTempo} is convex in $\bf p$, and its minimum with respect to $\bf p$ can be obtained by iterating the following closed-form update rule until convergence:
\begin{align}
    p_{l}^{k+1} &\leftarrow\sqrt{\frac{\hat{g}_{l}^{k}}{\hat{z}_{l}^{k}}}\;\text{for}\;l=1,\hdots,L \;\label{eq:BToeplitzUpdate}, \text{where} \\ 
    \hat{\bf g} &:=\mathrm{diag}({\bf P}^{k}{\bf Q}^{H}({\bf B}^{k})^{-1}{\bf M}_{\mathrm{time}}^{k}({\bf B}^{k})^{-1}{\bf Q}{\bf P}^{k}) \label{eq:diagI}  \\
    \hat{\bf z} &:=\mathrm{diag}({\bf Q}^{H}({\bf B}^{k})^{-1}{\bf Q})\;. 
    \label{eq:diagII} 
\end{align}
\end{theo}
\begin{proof}
A detailed proof is provided in Appendix~\ref{appendix:Time-Toeplitz}.
\end{proof}

\subsection{Efficient computation of the posterior}
Given estimates of the spatial and temporal covariance matrices, we can efficiently compute the posterior mean by exploiting their intrinsic diagonal structure. 
\begin{theo}
\label{theo:eff-post}
Given the diagonalization ${\bf B}={\bf Q}{\bf P}{\bf Q}^{H}$ of the temporal correlation matrix and the eigenvalue decomposition ${\bf L}\bm{\Gamma}{\bf L}^{\top}= {\bf U}_{\bf x}{\bf D}_{\bf x}{\bf U}_{\bf x}^{\top}$ of ${\bf L}\bm{\Gamma}{\bf L}^{\top}$, where ${\bf D}_{\bf x}=\mathrm{diag}(d_1,\hdots,d_M)$, the posterior mean is efficiently computed as 
\begin{align}
    \bar{\bf x}_{g} & =\left(\bm{\Gamma}\otimes{\bf B}\right){\bf D}^{\top}\Tilde{\bm \Sigma}_{\bf y}^{-1}{\bf y}_{g}= \tr\left({\bf Q}{\bf P} \left({\bm \Pi} \odot{\bf Q}^{H}{\bf Y}_{g}^{\top}{\bf U}_{\bf x}\right)\left({\bf U}_{{\bf x}}^{\top}{\bf L}\bm{\Gamma}^{\top}\right) \right) 
    \label{eq:efficientPosterior}\;,
\end{align}
where $\odot$ denotes the Hadamard product between the corresponding elements of two matrices of equal size. In addition, ${\bm \Pi}$ is defined as follows: $[{\bm \Pi}]_{l,m}= \frac{1}{{\sigma}_m^2 + p_l d_m}\;\text{for}\; l=1,\hdots,L\;\text{and}\;\;m=1,\hdots,M$.
\end{theo}
\begin{proof}
A detailed proof is provided in Appendix \ref{appendix:eff-post}. 
\end{proof}

The resulting algorithm, obtained by iterating between update rules \eqref{eq:BToeplitzUpdate}, \eqref{eq:HDiagonalUpdate} and \eqref{eq:efficientPosterior}, is called \emph{thin Dugh} (as opposed to \emph{full Dugh} introduced above).

\section{Simulations}
\label{sec:simulation}
\vspace{-3mm}
We present two sets of experiments to assess the performance of our proposed methods. In the first experiment, we compare the reconstruction performance of the proposed Dugh algorithm variants to that of Champagne \citep{wipf2010robust} and two other competitive methods -- eLORETA \citep{pascual2007discrete} and S-FLEX \citep{haufe2011large} -- for a range of SNR levels, numbers of time samples, and orders of AR coefficients. In the second experiment, we test the impact of model violations on the temporal covariance estimation.
All experiments are performed using \texttt{Matlab} on a machine with a 2.50 GHz Intel(R) Xeon(R) Platinum 8160 CPU. \ali{The computational complexity of each method in terms of the average running time in units of seconds for 1000 iterations is as follows: \emph{Full Dugh}: 67.094s, \emph{Thin Dugh}: 62.289s, \emph{Champagne}: 1.533s, \emph{eLORETA}: 2.653s, and  \emph{S-FLEX}: 20.963s.}
The codes are publicly available at \href{https://github.com/AliHashemi-ai/Dugh-NeurIPS-2021}{\textcolor{blue}{https://github.com/AliHashemi-ai/Dugh-NeurIPS-2021}}.

\vspace{-1mm}
\subsection{Pseudo-EEG signal generation and benchmark comparison}
\vspace{-1mm}

We simulate a sparse set of $N_0 =3$ active sources placed at random locations on the cortex. To simulate the electrical neural activity, we sample time series of length $T \in \{10,20,50,100\}$ from a univariate linear autoregressive AR(P) process.
We use stable AR systems of order $P \in \{1,2,5,7\}$. 
The resulting source distribution is then projected to the EEG sensors, denoted by ${{\bf Y}^{\mathrm{signal}}}$, using a realistic lead field matrix, ${\bf L} \in \mathbb{R}^{58 \times 2004}$. We generate ${\bf L}$ using the New York Head model \citep{huang2016new} taking into account the realistic anatomy and electrical tissue conductivities of an average human head. Finally, we add Gaussian white noise to the sensor space signal.
%
%
Note that the simulated noise and source time courses do not share a similar temporal structure here -- sources are modeled with a univariate autoregressive AR(P) process while a temporally white Gaussian distribution is used for modeling noise. Thus, we could assess the robustness of our proposed method under violation of the model assumption that the temporal structure of sources and noise is similar.
The resulting noise matrix ${\bf E}=[{\bf e}(1),\dots,{\bf e}(T)]$ is first normalized and then added to the signal matrix ${\bf Y}^{\mathrm{signal}}$ as follows: ${\bf Y}={\bf Y}^{\mathrm{signal}}+\frac{(1-\alpha)\left\Vert {{\bf Y}^{\mathrm{signal}}}\right\Vert_{F}}{\alpha \left\Vert{\bf E}\right\Vert_{F}}{\bf E}$, where $\alpha$ determines the signal-to-noise ratio (SNR) in sensor space. Precisely, SNR is defined as $\mathrm{SNR} = 20\mathrm{log}_{10}\left(\nicefrac{\alpha}{1-\alpha}\right)$. In this experiment the following values of $\alpha$ are used: $\alpha \in $ \{0.55, 0.65, 0.7, 0.8\}, which correspond to the following SNRs: SNR $\in$ \{1.7, 5.4, 7.4, 12\}~(dB). {Interested readers can refer to Appendix~\ref{appendix-Simulation} and \citep{haufe2016simulation} for further details on the simulation framework.}
%
We quantify the performance of all algorithms using the \emph{earth mover's distance} (EMD) \citep{rubner2000emd,haufe2008combining} and the maximal correlation between the time courses of the simulated and the reconstructed sources (TCE). Each simulation is carried out 100 times using different instances of $\bf{X}$ and $\bf E$, and the mean and standard error of the mean (SEM) of each performance measure across repetitions is calculated. 


In order to investigate the impact of model violations on the temporal covariance estimation, we generate a random Gaussian source matrix, ${\bf X} \in \mathbb{R}^{2004 \times 30 \times G}$ representing the brain activity of $2004$ brain sources at $30$ time instances for different numbers of trials $G \in 10,20,30,40,50$. In all trials, sources are randomly sampled from a zero-mean normal distribution with spatio-temporal covariance matrix ${\bm \Gamma} \otimes {\bf B}$, where ${\bf B} \in \mathbb{R}^{30 \times 30}$ is either a full-structural random PSD matrix or a Toeplitz matrix with $\beta = 0.8$. Gaussian noise $\bf E$ sharing the same temporal covariance with the sources is added to the measurements ${\bf Y}={\bf L}{\bf X}$, so that the overall SNR is $0$~dB. We evaluate the accuracy of the temporal covariance reconstruction using Thin and full Dugh.
The performance is evaluated using two measures: Pearson correlation between the original and reconstructed temporal covariance matrices, $\bf B$ and $\hat{{\bf B}}$, denoted by $r({\bf B},\hat{\bf B})$, and the normalized mean squared error (NMSE) defined as: $\text{NMSE} = ||\hat{{\bf B}}-{\bf B}||_{F}^{2} / ||{\bf B}||_{F}^{2}$. The similarity error is defined as: $1- r({\bf B},\hat{\bf B})$. Note that NMSE measures the reconstruction at the true scale of the temporal covariance; while $r({\bf B},\hat{\bf B})$ is scale-invariant and hence only quantifies the overall structural similarity between simulated and estimated noise covariance matrices. 

\begin{figure}
\begin{minipage}[b]{0.85\linewidth}
  \centering
    \centerline{\includegraphics[width=8cm]{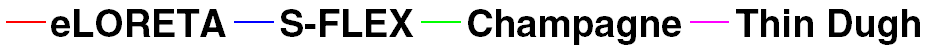}}
\end{minipage}
  \centering
  \centerline{\includegraphics[width=\textwidth,scale=0.5,keepaspectratio]{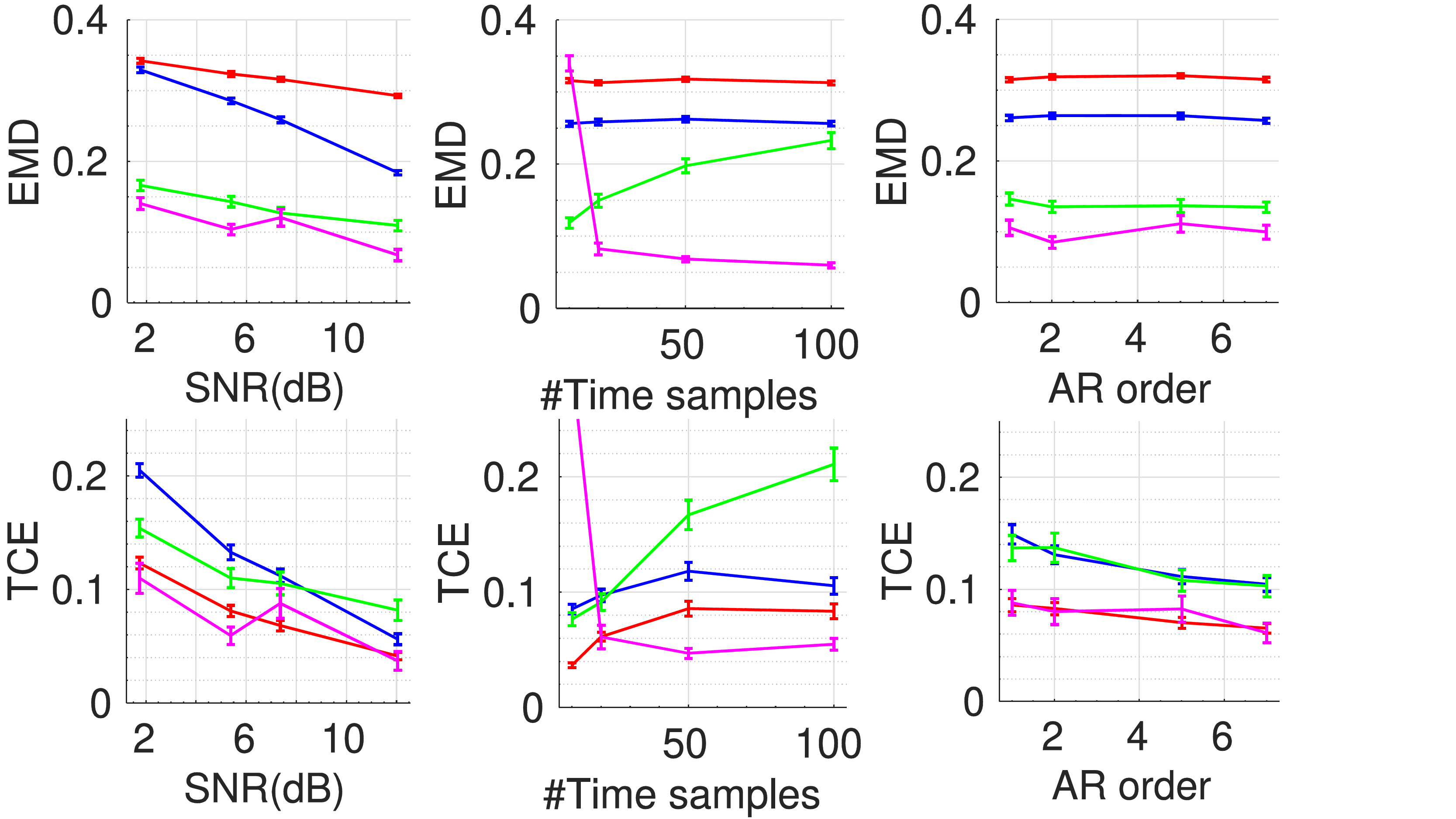}}
\caption{Source reconstruction performance (mean $\pm$ SEM) of the four different brain source imaging schemes, namely Thin Dugh, Champagne, eLORETA, and S-FLEX, for data generated by a realistic lead field matrix. Performance is assessed for different settings including a wide range of SNRs, different numbers of time samples, and different AR model orders. Performance is evaluated in terms of the earth mover's distance (EMD) and time-course correlation error (TCE) between each simulated source and the reconstructed source with highest maximum absolute correlation.}
\label{fig:metrics-source}
\vspace{-3mm}
\end{figure}

\begin{figure}
  \centering
  \centerline{\includegraphics[width=\textwidth,trim=0 5cm 0 0,clip]{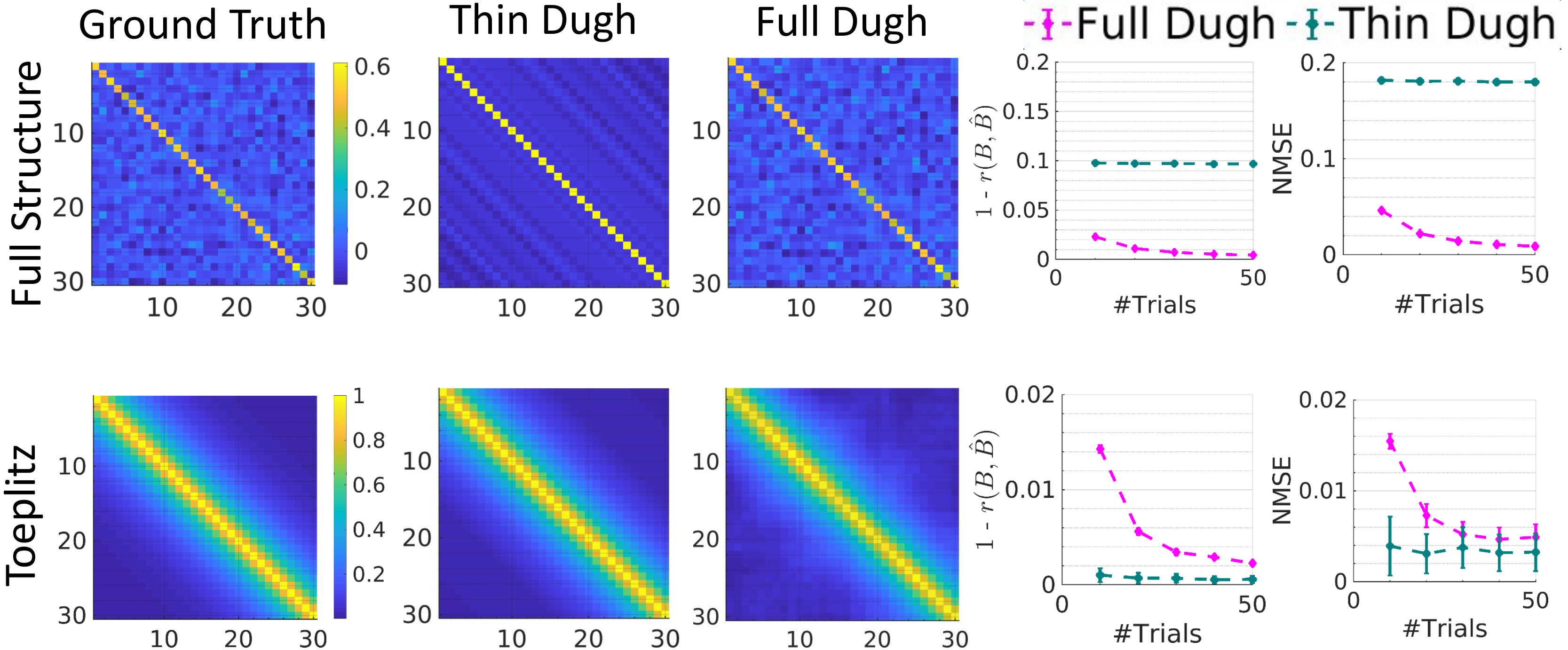}}
\caption{Accuracy of the temporal covariance matrix reconstruction incurred by two different temporal learning approaches, namely thin (cyan) and full (magenta) Dugh, assuming Toeplitz and full temporal covariance structure, respectively. The ground-truth temporal covariance matrix (first column) has full covariance structure in the first row, and Toeplitz structure in the second row. Performance is assessed in terms of the Pearson correlation between the entries of the original and reconstructed temporal covariance matrices, ${\bf B}$ and $\hat{{\bf B}}$, denoted by $r({\bf B},\hat{\bf B})$. Shown is the similarity error $1- r({\bf B},\hat{\bf B})$ (forth column). Further, the normalized mean squared error (NMSE) between ${\bf B}$ and $\hat{{\bf B}}$, defined as $\text{NMSE} = ||\hat{{\bf B}}-{\bf B}||_{F}^{2} /||{\bf B}||_{F}^{2}$ is reported (fifth column).}
\label{fig:metrics-temporal}
\vspace{-2mm}
\end{figure}

In Figure \ref{fig:metrics-source}, we show the source reconstruction performance (mean $\pm$ SEM) of the four different brain source imaging schemes, namely thin Dugh, Champagne, eLORETA and S-FLEX. We notice that Dugh achieves superior performance in terms of EMD metric, whereas it is competitive in terms of TCE. Note that since thin Dugh incorporates the temporal structure of the sources into the inference scheme, its performance with respect to EMD and TCE can be significantly improved by increasing the number of time samples.
Figure~\ref{fig:metrics-temporal} demonstrates the estimated temporal covariance matrix obtained from our two proposed spatio-temporal learning schemes, namely thin and full Dugh, indicated by cyan and magenta colors, respectively. The upper panel illustrates the reconstruction results for a setting where the ground-truth temporal covariance matrix has full structure, while the lower panel shows a case with Toeplitz temporal covariance matrix structure. It can be seen that \emph{full Dugh} can better capture the overall structure of ground truth full-structure temporal covariance as evidenced by lower NMSE and similarity errors compared to \emph{thin Dugh} that is only able to recover a Toeplitz matrix. As can be expected, the behavior is reversed in the lower-panel of Figure~\ref{fig:metrics-temporal}, where the ground truth temporal covariance matrix is indeed Toeplitz. Full Dugh, however, still provides reasonable performance in terms of NMSE and similarity error, even though it estimates a full-structure temporal covariance matrix. Finally, the performance of both learning schemes are significantly improved by increasing the number of trials. 
\vspace{-2mm}
\section{Analysis of real MEG and EEG data recordings}
\label{sec:real-data-analysis}
\vspace{-2mm}
\ali{We now demonstrate the performance of the novel algorithms on two MEG datasets (see Figure~\ref{fig:AEF} and \ref{fig:VEF}) and one EEG dataset (see Appendix \ref{appendix:real-data}). All participants provided informed written consent prior to study participation and received monetary compensation for their participation. The studies were approved by the University of California, San Francisco Committee on Human Research. The MEG datasets from two female subjects comprised five trials of visual or auditory stimulus presentation, where the goal was to reconstruct cortical activity reflecting auditory and visual processing. No prior work has shown success in reconstruction in such extreme low SNR data.} 

\ali{Figure~\ref{fig:AEF} shows the reconstructed sources for auditory evoked fields (AEF) from a representative subject using eLORETA, MCE, thin and full Dugh. In this case, we tested the reconstruction performance of all algorithms with the number of trials limited to 5. As Figure~\ref{fig:AEF} demonstrates, the reconstructions of thin and full Dugh both show focal sources at the expected locations of the auditory cortex. Limiting the number of trials to as few as 5 does not negatively influence the reconstruction result of Dugh methods, while it severely affects the reconstruction performance of competing methods. For the visual evoked field (VEF) data in Figure~\ref{fig:VEF}, thin Dugh was able to reconstruct two sources in the visual cortex with their corresponding time-courses demonstrating an early peak at~100 ms and a later peak around~200 ms (similar performance was observed for full Dugh).}
\begin{figure}
  \centering
  \centerline{\includegraphics[width=\textwidth,trim=0 17cm 0 0,clip]{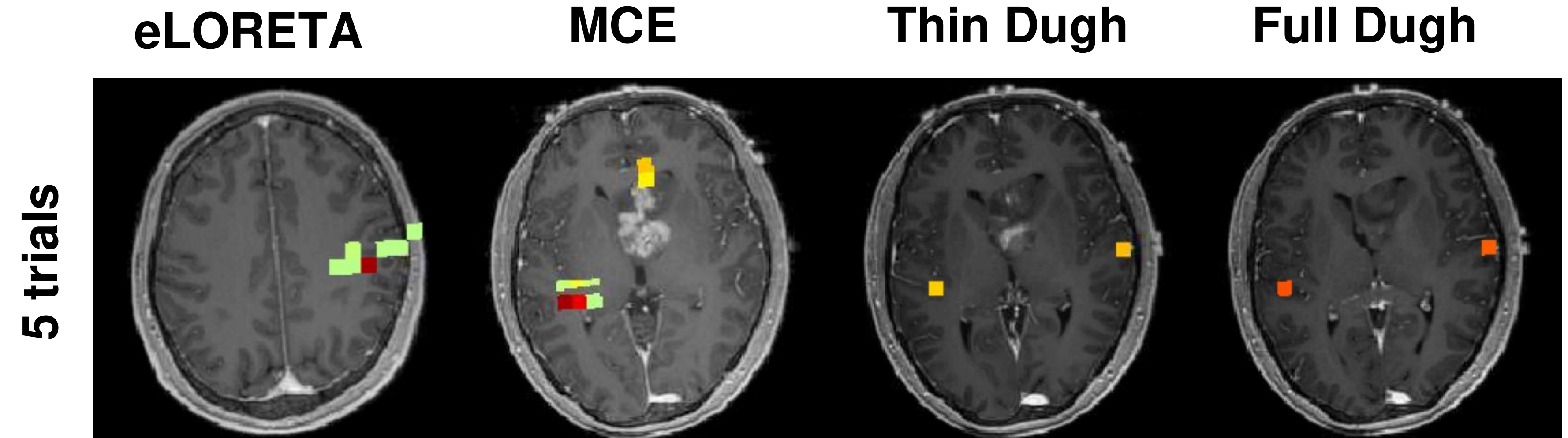}}
\caption{Analysis of auditory evoked fields (AEF) of one representative subject using eLORETA, MCE, thin and full Dugh. As it can be seen, thin and full Dugh can correctly localize bilateral auditory activity to Heschl's gyrus, which is the characteristic location of the primary auditory cortex, with as few as 5 trials. In this challenging setting, all competing methods show inferior performance.}
\label{fig:AEF}
\vspace{-2mm}
\end{figure}
\begin{figure}
  \centering
  \centerline{\includegraphics[width=\textwidth]{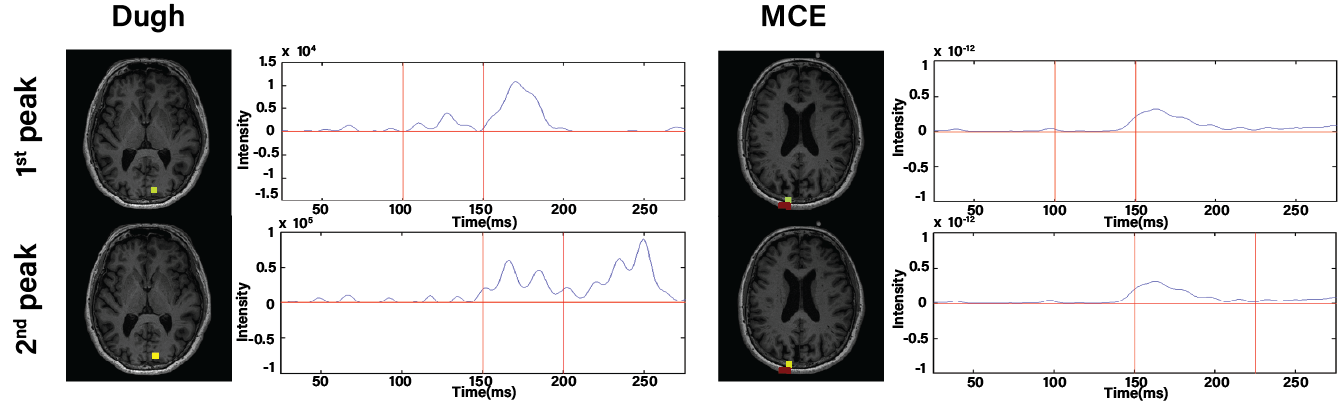}}
\caption{Spatial maps and corresponding time-courses reconstructed from five trials of visual stimulus presentation using Dugh and MCE. Distinct activity in visual cortex was reconstructed.}
\label{fig:VEF}
\vspace{-6mm}
\end{figure}
%

\section{Discussion}
\label{sec:discussion}
Inverse modeling is a challenging problem that can be cast as a multi-task regression problem. Machine learning-based methods have contributed by systematically incorporating prior information into such models (e.g.~\citep{wipf2009unified,wu2016bayesian,mccann2017convolutional,arridge2019solving,bubba2019learning}). While, so far, explicit models of spatio-temporal dynamics are rare, we contribute in this paper by deriving efficient inference algorithms for regression models with spatio-temporal dynamics in model parameters and noise. Specifically, we employ separable Gaussian distributions using Kronecker products of temporal and spatial covariance matrices. We assume sparsity and independent activations in the spatial domain, while the temporal covariance is modeled to have either full or Toeplitz structure. The proposed Dugh framework is encompassing efficient optimization algorithms for jointly estimating these distributions within a hierarchical Bayesian inference and MM optimization framework. Interestingly, we could theoretically prove convergence for the proposed update rules yielding estimates of the spatial and temporal covariances. In careful simulation studies, we have demonstrated that the inclusion of both spatial and temporal model parameters in the model indeed leads to significantly improved reconstruction performance. Finally, the utility of our algorithms is showcased in challenging real-world applications by reconstructing expected sources from real M/EEG data based on very few experimental trials.

The M/EEG BSI problem has been previously framed as a multi-task regression problem in the context of Type-I learning methods  \citep{massias2018generalized,bertrand2019handling}. \citet{bertrand2019handling} proposed a method to extend the group Lasso class of algorithms to the multi-task learning case. Their method shows promising results for large numbers of trials but is not competitive for smaller sample sizes, presumably due to not modeling the temporal structure of the sources. In contrast, Dugh is a Type-II method that learns the full spatio-temporal prior source distribution as part of the model fitting. Note that the assumption of spatially independent noise made here can be readily dropped, as in  \citep{hashemi2021joint}. A number of alternative approaches have been proposed to estimate the spatio-temporal correlation structure of the \emph{noise}
\citep{huizenga2002spatiotemporal,de2002estimating,bijma2003mathematical,de2004maximum,jun2006spatiotemporal}. These works, however, do not estimate the noise characteristics as part of the source reconstruction problem on the same data but require separate noise recordings. Our proposed algorithm substantially differs in this respect, as it learns the full noise covariance jointly with the brain source distribution. This joint estimation perspective is very much in line with the end-to-end learning philosophy as opposed to a step-wise independent estimation process that can give rise to error accumulation. 

With respect to the limits of our Dugh framework, we would like to note that using the same temporal correlation prior for noise and sources is a potentially restricting assumption in our modeling. It is made here to achieve tractable inference due to the fact that the spatio-temporal statistical model covariance matrix can be formulated in a separable form, i.e., $\Tilde{\bm \Sigma}_{\bf y} = {\bm \Sigma}_{\bf y} \otimes {\bf B}$. Although we have demonstrated empirically that the reconstruction results of our proposed learning schemes are fairly robust against violating this assumption, this constraint may be further relaxed by exploiting eigenvalue decomposition techniques presented in \citep{rakitsch2013all,wu2018learning}. Another potentially limiting assumption in our model is to assume Gaussian distributions for the sources and noise. Although Gaussian priors are commonly justified, the extension of our framework to heavy-tailed noise distributions, which are more robust to outliers, is a direction of future work. \ali{Regarding the societal impact of this work, we note that we intend to solve inverse problems that have non-unique solutions, which depend on explicit or implicit assumptions of priors. These have applications in basic, clinical, and translational neuroscience imaging studies. Our use of a hierarchical empirical Bayesian framework allows for explicit specifications of the priors in our model that are learned from data, and a more clear interpretation of the reconstructions with reduced bias. Nevertheless, in our algorithms, we assume sparse spatial priors, and in scenarios where this assumption embodied in the priors is incorrect, the resulting reconstructions will be inaccurate. Users and neuroscientists will need to be cognizant of these issues.} 

\label{sec:conclusion}
In conclusion, we could derive novel flexible hierarchical Bayesian algorithms for multi-task regression problems with spatio-temporal covariances. Incorporating prior knowledge, namely, constraining the solutions to Riemannian manifolds, and using ideas of geodesic convexity, circulant embeddings, and majorization-minimization, we derive inference update rules and prove convergence guarantees. The proposed Dugh algorithms show robust and competitive performance both on synthetic and real neural data from M/EEG recordings and thus contribute to a well-founded solution to complex inverse problems in neuroscience.

\section*{Broader Impact}
In this paper, we focused on sparse multi-task linear regression within the hierarchical Bayesian regression framework and its application in EEG/MEG brain source imaging. Our algorithm, however, is suitable for a wider range of applications. The same concepts used here for full-structural spatio-temporal covariance learning could be employed in other contexts where hyperparameters like kernel widths in Gaussian process regression \citep{wu2019dependent} or dictionary elements in the dictionary learning problem \citep{dikmen2012maximum} need to be inferred from data. Our proposed method, Dugh, may also prove useful for practical scenarios in which model residuals and signals are expected to be correlated, e.g., probabilistic canonical correlation analysis (CCA) \citep{bach2005probabilistic}, spectral independent component analysis (ICA) \citep{ablin2020spectral}, direction of arrival (DoA) and channel estimation in massive Multiple Input Multiple Output (MIMO) systems \citep{prasad2015joint,gerstoft2016multisnapshot,haghighatshoar2017massive}, robust portfolio optimization in finance \citep{feng2016signal}, covariance matching and estimation \citep{werner2008estimation,tsiligkaridis2013covariance,zoubir2018robust,benfenati2020proximal,ollila2020shrinking}, graph learning \citep{kumar2020unified}, thermal field reconstruction \citep{hashemi2016efficient,flinth2017thermal,flinth2018approximate}, and brain functional imaging \citep{wei2020bayesian}.

\begin{ack}
This result is part of a project that has received funding from the European Research Council (ERC) under the European Union’s Horizon 2020 research and innovation programme (Grant agreement No. 758985).
AH acknowledges scholarship support from the Machine Learning/Intelligent Data Analysis research group at Technische Universit{\"a}t Berlin. He further wishes to thank the  Charit{\'e} -- Universit{\"a}tsmedizin Berlin, the Berlin Mathematical School (BMS), and the Berlin Mathematics Research Center MATH+ for partial support. CC was supported by the National Natural Science Foundation of China under Grant 62007013. KRM was funded by the German Ministry for Education and Research as BIFOLD -- Berlin Institute for the Foundations of Learning and Data (ref.\ 01IS18025A and ref.\ 01IS18037A), and the German Research Foundation (DFG) as Math+: Berlin Mathematics Research Center  (EXC 2046/1, project-ID: 390685689),
Institute of Information \& Communications Technology Planning \& Evaluation (IITP) grants funded by the Korea Government (No. 2019-0-00079,  Artificial Intelligence Graduate School Program, Korea University). SSN was funded in part by National Institutes of Health grants (R01DC004855, R01EB022717, R01DC176960, R01DC010145, R01NS100440, R01AG062196, and R01DC013979), University of California MRPI MRP-17–454755, the US Department of Defense grant (W81XWH-13-1-0494).
\end{ack}



\bibliography{ref}
\bibliographystyle{unsrtnat} 


\newpage
\appendix

\begin{center}
\textbf{\Large Supplementary Material: Efficient hierarchical Bayesian inference for spatio-temporal regression models in neuroimaging}
\end{center}


\section*{Summary of the proposed algorithms and derived update rules}
We proposed two algorithms in the main paper, namely full and thin Dugh, which are summarized in Algorithm~\ref{alg:Full-Dugh} and Algorithm~\ref{alg:Thin-Dugh}, respectively. 
\begin{algorithm}
    \SetAlgoLined
    \SetKwInOut{Input}{Input}
    \Input{The lead field matrix ${\bf L} \in \mathbb{R}^{M\times N}$ and $G$ trials of measurement vectors $\{{\bf Y}_{g}\}_{g=1}^{G}$, where ${\bf Y}_{g} \in \mathbb{R}^{M\times T}$.}
    \KwResult{Estimates of the source and noise variances ${\bf h} = [\gamma_{1},\dots,\gamma_{N}, \sigma^2_{1},\dots,\sigma^2_{M}]^\top$, the temporal covariance $\bf B$, and the posterior mean $\{\bar{\bf x}_{g}\}_{g=1}^{G}$ and covariance ${\bm \Sigma}_{\bf x}$ of the sources.} 
        \nl Choose a random initial value for $\bf B$ as well as ${\bf h} = [\gamma_{1},\dots,\gamma_{N}, \sigma^2_{1},\dots,\sigma^2_{M}]^\top$, and construct ${\bf H} =\mathrm{diag}({\bf h})$ and ${\bm \Gamma} =\mathrm{diag}([\gamma_{1},\dots,\gamma_{N}]^\top)$.
        
        \nl Construct the augmented lead field ${\bm \Phi}=[{\bf L},{\bf I}_M]$.
        
        \nl Calculate the lead field ${\bf D}={\bf L}\otimes {\bf I}_{T}$ for vectorized sources.
        
       \nl Calculate the prior spatio-temporal covariance for the sources as ${\bm \Sigma}_{0}={\bm \Gamma} \otimes {\bf B}$. 
                   
        \nl Calculate the spatial statistical covariance ${\bm \Sigma}_{\bf y}={\bm \Phi}{\bf H}{\bm \Phi}^{\top}$. 
        
        \nl Calculate the spatio-temporal statistical covariance $\Tilde{\bm \Sigma}_{\bf y} = {\bm \Sigma}_{\bf y} \otimes {\bf B}$.
        
        \nl Initialize {$k \leftarrow 1$}
        
        \Repeat{stopping condition is satisfied: $\left\Vert \bar{\bf x}^{k+1} - \bar{\bf x}^{k} \right\Vert^{2}_{2} \leq \epsilon$ or $k = k_{\text{max}}$}{
            \nl Calculate the posterior mean as $\bar{\bf x}_{g}={\bm \Sigma}_{0}{\bf D}^{\top}\Tilde{\bm \Sigma}_{\bf y}^{-1}{\bf y}_{g}$, for $g=1,\ldots,G$, where ${\bf y}_{g}=\text{vec}\left({\bf Y}_{g}^\top \right)\in\mathbb{R}^{MT\times1}$. 
            
            \nl Calculate ${\bf M}_{\mathrm{time}}^{k}$ based on Eq.~\eqref{eq:TempSampleCov}, and update ${\bf B}$ based on Eq.~\eqref{eq:BGeodesicUpdate} according to the Riemannian update on the manifold of P.D. matrices.
            
            \nl {Calculate ${\bf M}_{\mathrm{SN}}^{k}$ based on Eq.~\eqref{eq:SpaceSampleCov}, and update ${\bf H}$ based on Eq.~\eqref{eq:HDiagonalUpdate}.}
            
            \nl {$k \leftarrow k + 1$}
            }      
           
           \nl Calculate the posterior covariance as ${\bm \Sigma}_{\bf x}= {\bm \Sigma}_{0}-{\bm \Sigma}_{0}{\bf D}^{\top}\Tilde{\bm \Sigma}_{\bf y}^{-1}{\bf D}{\bm \Sigma}_{0}$.

    \caption{Full Dugh} 
    \label{alg:Full-Dugh}
\end{algorithm}
\vspace{-10mm}
\begin{algorithm}
    \SetAlgoLined
    \SetKwInOut{Input}{Input}
    \Input{The lead field matrix ${\bf L} \in \mathbb{R}^{M\times N}$, and $G$ trials of measurement vectors $\{{\bf Y}_{g}\}_{g=1}^{G}$, where ${\bf Y}_{g} \in \mathbb{R}^{M\times T}$.}
    \KwResult{Estimates of the source and noise variances ${\bf h} = [\gamma_{1},\dots,\gamma_{N}, \sigma^2_{1},\dots,\sigma^2_{M}]^\top$, the temporal covariance $\bf B$, and the posterior mean $\{\bar{\bf x}_{g}\}_{g=1}^{G}$.} 
    
        \nl Choose a random initial value for $\bf p$ as well as $\bf h$, and construct ${\bf H} =\mathrm{diag}({\bf h})$ and ${\bf P} =\mathrm{diag}({\bf p})$.
        
        \nl Construct ${\bf B}={\bf Q}{\bf P}{\bf Q}^{H}$, where $\mathbf{Q} = [{\bf I}_{M},{\bf 0}]{\bf F}_{L}$ with $L =2T +1$ and ${\bf F}_{L}$ as DFT.
        
        \nl Construct the augmented lead field ${\bm \Phi}:=[{\bf L},{\bf I}_M]$.
        
        \nl Calculate the lead field ${\bf D}={\bf L}\otimes {\bf I}_{T}$ for vectorized sources.
        
        \nl Calculate the statistical covariance ${\bm \Sigma}_{\bf y}={\bm \Phi}{\bf H}{\bm \Phi}^{\top}$. 
        
        \nl Calculate the statistical covariance ${\bm \Sigma}_{\bf y}={\bm \Phi}{\bf H}{\bm \Phi}^{\top}$. 
        
        \nl Calculate the spatio-temporal statistical covariance $\Tilde{\bm \Sigma}_{\bf y} = {\bm \Sigma}_{\bf y} \otimes {\bf B}$.
        
        \nl Initialize  {$k \leftarrow 1$}
        
        \Repeat{stopping condition is satisfied: $\left\Vert \bar{\bf x}^{k+1} - \bar{\bf x}^{k} \right\Vert^{2}_{2} \leq \epsilon$ or $k = k_{\text{max}}$}{
            \nl Calculate the posterior mean efficiently based on Eq.~\eqref{eq:efficientPosterior} as $\bar{\bf x}_{g} = \tr\left({\bf Q}{\bf P} \left({\bm \Pi} \odot{\bf Q}^{H}{\bf Y}_{g}^{\top}{\bf U}_{\bf x}\right)\left({\bf U}_{{\bf x}}^{\top}{\bf L}\bm{\Gamma}^{\top}\right) \right)$, where ${\bf L}\bm{\Gamma}{\bf L}^{\top}= {\bf U}_{\bf x}{\bf D}_{\bf x}{\bf U}_{\bf x}^{\top}$ and $[{\bm \Pi}]_{l,m}= \frac{1}{{\sigma}_m^2 + p_l d_m}\;\text{for}\; l=1,\hdots,L\;\text{and}\;\;m=1,\hdots,M$.
            
            \nl Calculate ${\bf M}_{\mathrm{time}}^{k}$ based on Eq.~\eqref{eq:TempSampleCov}, and update ${\bf B}$ based on Eq.~\eqref{eq:BToeplitzUpdate} according to Riemannian update for Toeplitz matrices using circulant embedding.
            
            \nl {Calculate ${\bf M}_{\mathrm{SN}}^{k}$ based on Eq.~\eqref{eq:SpaceSampleCov}, and update ${\bf H}$ based on Eq.~\eqref{eq:HDiagonalUpdate}.}
            
            \nl {$k \leftarrow k + 1$}
            }     
            
           \nl Calculate the posterior covariance as ${\bm \Sigma}_{\bf x}= {\bm \Sigma}_{0}-{\bm \Sigma}_{0}{\bf D}^{\top}\Tilde{\bm \Sigma}_{\bf y}^{-1}{\bf D}{\bm \Sigma}_{0}$.
            
    \caption{Thin Dugh} 
    \label{alg:Thin-Dugh}
\end{algorithm}

\newpage
\section{Derivation of Type-II Bayesian cost function for full-structural spatio-temporal models}
\label{appendix:spatio-temp-cost}


In this section, we provide a detailed derivation of Type-II Bayesian learning for full-structural spatio-temporal models. To this end, we first briefly explain the $\emph{multiple measurement vector}$ (MMV) model and then formulate Type-II Bayesian learning with full-structural spatio-temporal covariance structure for this setting. Note that, to simplify the problem, we first present the derivations of the MMV model only for a single trial. We later extend this simplified setting to the multi-trials case. 



\subsection{Multiple measurement vector (MMV) model}

In M/EEG brain source imaging, a sequence of measurement vectors are often available. Thus, the following $\emph{multiple measurement vector (MMV)}$ 
model can be formulated1:
\begin{align*}
    {\bf Y}={\bf LX}+{\bf E}\;,
\end{align*}
where ${\bf Y}=[{\bf y}(1),\dots,{\bf y}(T)]\in\mathbb{R}^{M\times T}$ consists of $T$ measurement vectors for a sequence of $T$ time samples. ${\bf X}=[{\bf x}(1),\dots,{\bf x}(T)]\in\mathbb{R}^{N\times T}$ is the desired solution matrix (the amplitude of $N$ brain sources during $T$ time samples in our setting), and ${\bf E}$ is an unknown noise matrix. A key assumption in the MMV model is that the support (i.e., the indices of the nonzero entries) of every column in ${\bf X}$ is identical (referred to as the $\emph{common sparsity assumption}$ in the literature). 
The number of nonzero rows in ${\bf X}$ needs to be below a threshold to ensure unique and global solution. This implies that ${\bf X}$ has only a small number of non-zero rows. It has been shown that the recovery of the support can be greatly improved by increasing the number of measurements \citep{eldar2009robust,jin2013support,pal2014pushing}.

\subsection{Type-II Bayesian cost function for full-structural spatio-temporal models}

To exploit temporal correlations between measurements, we first assume that the voxels are mutually independent. Given the column vector $\bm \gamma=[\gamma_{1},\dots,\gamma_{N}]^{\top}$ and a Gaussian probability density for each brain source, the prior distribution with time correlation is modeled as follows: 
\begin{align}
    p({\bf X}_{i}|\gamma_{i},{\bf B})\sim\mathcal{N}(0,\gamma_{i}{\bf B}),\qquad i=1,\dots,N\;.
\end{align}
${\bf X}_{i}$ denotes the $i$-th row of source matrix ${\bf X}$ and models the probability distribution of the $i$-th brain source. Note that $\gamma_{i}$ is a non-negative hyper-parameter that controls the row sparsity of ${\bf X}$; i.e., the values of source ${\bf X}_{i}$ become all zero if $\gamma_{i}=0$. Finally, ${\bf B}$ is a positive definite matrix that captures the time correlation structure, which is assumed to be shared across all sources. The goal is to obtain the prior distribution of sources, $p({\bf X}|{\bm \gamma},{\bf B})$, by estimating the hyper-parameters, $\{{\bm \gamma},{\bf B}\}$.
Next, we reformulate the joint MMV model of all sources using vectorization of matrices and Kronecker product operations:
\begin{align*}
    {\bf y}={\bf D}{\bf x}+{\bf e}\;,
\end{align*}
where ${\bf y}=\text{vec}\left({\bf Y}^\top \right)\in\mathbb{R}^{MT\times1}$, ${\bf x}=\text{vec}\left({\bf X}^\top\right)\in\mathbb{R}^{NT\times1}$, ${\bf e}=\text{vec}\left({\bf E}^\top\right)\in\mathbb{R}^{MT\times1}$ and ${\bf D}$=${\bf L}\otimes {\bf I}_{T}$.

The prior distribution of ${\bf x}$ is given as
\begin{align*}
    p({\bf x}|{\bm \gamma},{\bf B})\sim\mathcal{N}(0,{\bm \Sigma}_{0})
\end{align*}
where ${\bm \Sigma}_{0}$ is defined as
\begin{align*}
    {\bm \Sigma}_{0}=\left[\begin{array}{ccc}
    \gamma_{1} {\bf B}\\
     & \ddots\\
     &  & \gamma_{N} {\bf B}
    \end{array}\right]={\bm \Gamma} \otimes {\bf B} \;,
\end{align*}
in which $\bm \Gamma=\text{diag}(\bm \gamma)=\text{diag}(\gamma_{1},\dots,\gamma_{N})$. 

Similarly, we may assume zero-mean Gaussian noise with covariance ${\bm \Sigma}_{\mathrm{\bf e}}={\bm \Lambda} \otimes {\bm \Upsilon}$, where ${\bf e}\sim \mathcal{N}(0,{\bm \Sigma}_{\mathrm{\bf e}})$, and ${\bm \Lambda}$ and ${\bm \Upsilon}$ denote the spatial and temporal noise covariance matrices, respectively. Here, we use the same prior for the temporal structure of noise and sources, i.e., ${\bm \Upsilon}={\bf B}$.

The parameters of the spatio-temporal Type-II model are the unknown source, noise and temporal covariance matrices, i.e., ${\bm \Gamma}$, ${\bm \Lambda}$, and ${\bf B}$. The unknown parameters ${\bm \Gamma}$, ${\bm \Lambda}$, and ${\bf B}$ are optimized based on the current estimates of the source, noise and temporal covariances in an alternating iterative process. Given initial estimates of ${\bm \Gamma}$, ${\bm \Lambda}$, and ${\bf B}$, the posterior distribution of the sources is a Gaussian of the form $p({\bf x}|{\bf y},{\bm \Gamma},{\bm \Lambda},{\bf B}) \sim \mathcal{N}(\bar{\bf x},{\bm \Sigma}_{\bf x})$, whose mean and covariance are obtained as follows:
\begin{align}
    \bar{\bf x} &={\bm \Sigma}_{0}{\bf D}^{\top}({\bm \Lambda} \otimes {\bf B} + {\bf D}{\bm \Sigma}_{0}{\bf D}^{\top})^{-1}{\bf y}={\bm \Sigma}_{0}{\bf D}^{\top}\Tilde{\bm \Sigma}_{\bf y}^{-1}{\bf y}\;,
    \\
    {\bm \Sigma}_{\bf x} &= {\bm \Sigma}_{0}-{\bm \Sigma}_{0}{\bf D}^{\top}\Tilde{\bm \Sigma}_{\bf y}^{-1}{\bf D}{\bm \Sigma}_{0}
    \;,
\end{align}
where ${\bm \Sigma}_{\bf y}={\bf L}{\bm \Gamma}{\bf L}^{\top}+{\bm \Lambda}$, and where $\Tilde{\bm \Sigma}_{\bf y} = {\bm \Sigma}_{\bf y} \otimes {\bf B}$ denotes the spatio-temporal variant of statistical model covariance matrix. The estimated posterior parameters $\bar{\bf x}$ and ${\bm \Sigma}_{\bf x}$ are then in turn used to update ${\bm \Gamma}$, ${\bm \Lambda}$, and $\bf B$ as the minimizers of the negative log of the marginal likelihood $p({\bf Y}|{\bm \Gamma},{\bm \Lambda},{\bf B})$, which is given by
\begin{align}
      \mathcal{L}_{\text{kron}}({\bm \Gamma},{\bm \Lambda},{\bf B}) &= \mathrm{log}|\Tilde{\bm \Sigma}_{\bf y}|+\tr\left({\bf y}^{\top} \Tilde{\bm \Sigma}_{\bf y}^{-1}{\bf y}\right) \;.
\end{align}
Using the same temporal covariance prior for noise and sources, i.e., ${\bm \Upsilon}={\bf B}$, the statistical model covariance matrix, $\Tilde{\bm \Sigma}_{\bf y}$, can be written as:
\begin{align}
    \Tilde{\bm \Sigma}_{\bf y} &= {\bm \Lambda} \otimes {\bm \Upsilon} + ({\bf D}{\bm \Sigma}_0 {\bf D}^{\top}) \nonumber = {\bm \Lambda} \otimes {\bm \Upsilon} + \left( \left({\bf L} \otimes {\bf I}^{\top}\right)\left({\bm \Gamma}\otimes {\bf B}\right)\left({\bf L}\otimes {\bf I}^{\top}\right)^{\top}\right) \nonumber  \\
     & = {\bm \Lambda} \otimes {\bm \Upsilon}+ \left({\bf L}{\bm \Gamma} {\bf L}^{\top} \otimes {\bf B} \right) \stackrel{({\bm \Upsilon}={\bf B})}{=}({\bm \Lambda}+{\bf L}{\bm \Gamma} {\bf L}^{\top}) \otimes {\bf B} \nonumber \\ 
     &= {\bm \Sigma}_{\bf y} \otimes {\bf B} \;,
     \;.\label{eq:Approximation}
\end{align}
which leads to the following spatio-temporal Type-II Bayesian learning cost function:
\begin{align}
      \mathcal{L}_{\text{kron}}({\bm \Gamma},{\bm \Lambda},{\bf B}) &= \mathrm{log}|\Tilde{\bm \Sigma}_{\bf y}|+\tr\left({\bf y}^{\top} \Tilde{\bm \Sigma}_{\bf y}^{-1}{\bf y}\right) \nonumber  \\
      &= \mathrm{log}|{\bm \Sigma}_{\bf y} \otimes {\bf B}|+\tr \left({\bf y}^{\top} \left({\bm \Sigma}_{\bf y} \otimes {\bf B}\right)^{-1} {\bf y}\right) \nonumber \\
      &= \mathrm{log}\left(|{\bm \Sigma}_{\bf y}|^{T} |{\bf B}|^M \right) + \tr\left({\bf y}^{\top} \left({\bm \Sigma}_{\bf y} \otimes {\bf B} \right)^{-1} {\bf y}\right)\;.
      \label{eq:appendix-vector-Type-II-loss}
\end{align}
Here, we assume the presence of $G$ sample blocks ${\bf Y}_{g} \in \mathbb{R}^{M\times T},\mathrm{for}\;g=1,\hdots,G$. These block samples can be obtained by segmenting a time series into smaller parts that are assumed to independent and identically distributed. These blocks may represent epochs, trials or experimentl tasks depending on the applications. $\mathcal{L}_{\text{kron}}({\bm \Gamma},{\bm \Lambda},{\bf B})$ can then be reformulated as 
\begin{align}
    \mathcal{L}_{\text{kron}}({\bm \Gamma},{\bm \Lambda},{\bf B}) &= T\log|{{\bm \Sigma}_{\bf y}}|+M\log|{\bf B}| +\frac{1}{G}\sum_{g=1}^{G}\text{tr}({{\bm \Sigma}_{\bf y}^{-1}}{\bf Y}_{g}{\bf B}^{-1}{\bf Y}_{g}^{\top}) 
\end{align}
by applying the following matrix equality to Eq.~\eqref{eq:appendix-vector-Type-II-loss}: 
\begin{equation*}
    \tr({\bf A}^{\top}{\bf B}{\bf C}{\bf D}^{\top})=\mathrm{vec}({\bf A})^{\top}({\bf D}\otimes {\bf B})\mathrm{vec}({\bf C}) \;.
\end{equation*}

\section{Proof of Theorem~\ref{Theo:Time-Surrogate}}
\label{appendix:Time-Surrogate}

\begin{proof}

We start by recalling $\mathcal{L}_{\text{kron}}({\bm \Gamma},{\bm \Lambda},{\bf B})$ in Eq.~\eqref{eq:MLKronokCost}:
\begin{align*}
    \mathcal{L}_{\text{kron}}({\bm \Gamma},{\bm \Lambda},{\bf B}) &= T\log|{{\bm \Sigma}_{\bf y}}|+M\log|{\bf B}| +\frac{1}{G}\sum_{g=1}^{G}\text{tr}({{\bm \Sigma}_{\bf y}^{-1}}{\bf Y}_{g}{\bf B}^{-1}{\bf Y}_{g}^{\top}).
\end{align*}

Let ${\bm \Sigma}_{\bf y}^{k}$, ${\bm \Gamma}^{k}$, and ${\bm \Lambda}^{k}$ be the values of statistical model covariance and the source and noise covariances at the $k$-th iteration, respectively. By ignoring terms that do not depend on ${\bf B}$, $\mathcal{L}_{\text{kron}}({\bm \Gamma},{\bm \Lambda},{\bf B})$ can be written as follows:
\begin{align}
    \mathcal{L}_{\text{kron}}^{\mathrm{time}}({\bm \Gamma}^{k},{\bm \Lambda}^{k},{\bf B}) 
    &= M\log|{\bf B}| +\frac{1}{G}\sum_{g=1}^{G}\tr\left(\left({\bm \Sigma}_{\bf y}^{k}\right)^{-1}{\bf Y}_{g}{\bf B}^{-1}{\bf Y}_{g}^{\top}\right) \nonumber \\ 
    &= \log|{\bf B}| +\frac{1}{MG}\sum_{g=1}^{G}\tr\left(\left({\bm \Sigma}_{\bf y}^{k}\right)^{-1}{\bf Y}_{g}{\bf B}^{-1}{\bf Y}_{g}^{\top}\right) \nonumber \\
    &= \log|{\bf B}|+\tr\left({\bf B}^{-1}\frac{1}{MG}\sum_{g=1}^{G}{\bf Y}_{g}^{\top}\left({\bm \Sigma}_{\bf y}^{k}\right)^{-1}{\bf Y}_{g}\right) \nonumber \\
    &= \log|{\bf B}| +\tr\left({\bf B}^{-1}{\bf M}_{\mathrm{time}}^{k}\right)\;,
\end{align}
where ${\bf M}_{\mathrm{time}}^{k}:=\frac{1}{MG}\sum_{g=1}^{G}{\bf Y}_{g}^{\top}\left({\bm \Sigma}_{\bf y}^{k}\right)^{-1}{\bf Y}_{g}$. 

By virtue of the concavity of the log-determinant function and its first order Taylor expansion around ${\bf B}^{k}$, the following inequality holds:
\begin{align}
    \mathcal{L}_{\text{kron}}^{\mathrm{time}}({\bm \Gamma}^{k},{\bm \Lambda}^{k},{\bf B}) 
    &= \log|{\bf B}|+\tr\left({\bf B}^{-1}{\bf M}_{\mathrm{time}}^{k}\right) \nonumber \\
    &\leq \log\left|{\bf B}^{k}\right| + \tr\left(\left({\bf B}^{k}\right)^{-1}\left({\bf B} -{\bf B}^{k} \right) \right) +\tr\left({\bf B}^{-1}{\bf M}_{\mathrm{time}}^{k}\right) \nonumber \\
    &= \log\left|{\bf B}^{k}\right| + \tr\left(\left({\bf B}^{k}\right)^{-1}{\bf B} \right)-\tr\left(\left({\bf B}^{k}\right)^{-1}{\bf B}^{k} \right) +\tr\left({\bf B}^{-1}{\bf M}_{\mathrm{time}}^{k}\right) \nonumber \\
    &= \tr\left(({\bf B}^{k})^{-1}{\bf B}\right)+\tr({\bf B}^{-1}{\bf M}_{\mathrm{time}}^{k}) + \text{const}\nonumber \\ 
    &= \mathcal{L}^{\mathrm{time}}_{\mathrm{conv}}({\bm \Gamma}^{k},{\bm \Lambda}^{k},{\bf B}) + \text{const}\;.
    \label{eq:Time-Surrogate-derivation}
\end{align}
Note that constant values in \eqref{eq:Time-Surrogate-derivation} do not depend on ${\bf B}$; hence, they can be ignored in the optimization procedure. Hence, we have shown that minimizing Eq.~\eqref{eq:MLKronokCost} with respect to ${\bf B}$ is equivalent to minimizing $\mathcal{L}^{\mathrm{time}}_{\mathrm{conv}}({\bm \Gamma}^{k},{\bm \Lambda}^{k},{\bf B})$, which concludes the proof.
\end{proof}

\section{Proof of Theorem~\ref{Theo:Time-Sol}}
\label{appendix:Time-Sol}
Before presenting the proof, the subsequent definitions and propositions are required: 
\begin{defi}[Geodesic path]\label{def:geodesic-path}
Let $\mathcal{M}$ be a Riemannian manifold, i.e., a differentiable manifold whose tangent space is endowed with an inner product that defines local Euclidean structures. Then, a geodesic between two points on $\mathcal{M}$, denoted by ${\bf p}_{0},{\bf p}_{1}\in\mathcal{M}$, is defined as the shortest connecting path between those two points along the manifold, ${\zeta}_{l}({{\bf p}_{0},{\bf p}_{1}})\in\mathcal{M}\;\text{for}\;l\in[0,1]$, where $l=0$ and $l=1$ defines the starting and end points of the path, respectively.
\end{defi}

\label{rem:Geo-path}
In the current context, $\zeta_l({\bf p}_{0},{\bf p}_{1})$ defines a geodesic curve on the P.D. manifold joining two P.D. matrices, ${\bf P}_0,{\bf P}_1 > 0$. The specific pair of matrices we will deal with is
$\{{\bf B}^{k},{\bf M}_{\mathrm{time}}^{k}\}$.
%
\begin{defi}[Geodesic on the P.D. manifold]\label{def:PSD-geo-path}
Geodesics on the manifold of P.D. matrices can be shown to form a cone within the embedding space. We denote this manifold by $\mathcal{S}_{++}$. Assume two P.D. matrices ${\bf P}_0,{\bf P}_1 \in \mathcal{S}_{++}$. Then, for $l\in[0,1]$, the geodesic curve joining ${\bf P}_0$ to ${\bf P}_1$ is defined as \citep[Chapter. 6]{bhatia2009positive}: 
\begin{equation}
    \xi_l({\bf P}_{0},{\bf P}_{1}) = ({\bf P}_0)^{\frac{1}{2}}\left(({\bf P}_0)^{\nicefrac{-1}{2}}{\bf P}_1({\bf P}_0)^{\nicefrac{-1}{2}}\right)^{l}({\bf P}_0)^{\frac{1}{2}}\; \quad l\in[0,1] \;.
    \label{eq:PSD-geodesic-path}        
\end{equation}
\end{defi}
\label{remark:geometric-mean}
Note that ${\bf P}_0$ and ${\bf P}_1$ are obtained as the starting and end points of the geodesic path by choosing $l=0$ and $l=1$, respectively. The midpoint of the geodesic, obtained by setting $l=\frac{1}{2}$, is called the \emph{geometric mean}. Note that, according to Definition~\ref{def:PSD-geo-path}, the following equality holds :
\begin{align}
    \xi_l({\bf B}_{0},{\bf B}_{1})^{-1} &= \left(({\bf B}_0)^{\nicefrac{1}{2}}\left(({\bf B}_0)^{\nicefrac{-1}{2}}{\bf B}_1({\bf B}_0)^{\nicefrac{-1}{2}}\right)^{l}({\bf B}_0)^{\nicefrac{1}{2}}\right)^{-1} \nonumber \\ &= \left(({\bf B}_0)^{\nicefrac{-1}{2}}\left(({\bf B}_0)^{\nicefrac{1}{2}}({\bf B}_1)^{-1}({\bf B}_0)^{\nicefrac{1}{2}}\right)^{l}({\bf B}_0)^{\nicefrac{-1}{2}}\right)
    = \xi_l({\bf B}_{0}^{-1},{\bf B}_{1}^{-1})\;.
    \label{eq:geodesic-PSD-inverse}
\end{align}
%
\begin{defi}[Geodesic convexity]\label{def:geodesic-convexity}
Let ${\bf p}_{0}$ and ${\bf p}_{1}$ be two arbitrary points on a subset $\ensuremath{\mathcal{\mathcal{A}}}$ of a Riemannian manifold $\ensuremath{\mathcal{M}}$. Then, a real-valued function$\ensuremath{f:\mathcal{\mathcal{A}\rightarrow\mathbb{R}}}$ with domain $\ensuremath{\mathcal{\mathcal{A}\subset M}}$ is called \emph{geodesic convex} (g-convex) if the following relation holds: 
\begin{equation}
    f\left(\zeta_l({\bf p}_{0},{\bf p}_{1})\right)\leq lf({\bf p}_{0})+(1-l)f({\bf p}_{1})\;,
    \label{eq:g-convex-condition}
\end{equation}
where $l\in[0,1]$ and $\zeta({\bf p}_{0},{\bf p}_{1})$ denotes the geodesic path connecting two points ${\bf p}_{0}$ and ${\bf p}_{1}$ as defined in Definition~\ref{def:geodesic-path}. 
Thus, in analogy to classical convexity, the function $f$ is g-convex if every geodesic $\zeta({\bf p}_{0},{\bf p}_{1})$ of $\mathcal{M}$ between ${\bf p}_{0},{\bf p}_{1}\in\mathcal{A}$, lies in the g-convex set $\mathcal{A}$. Note that the set $\mathcal{A} \subset\mathcal{M}$ is called g-convex, if any geodesics joining an arbitrary pair of points lies completely in $\mathcal{A}$.
\end{defi}
\begin{rem}
Note that g-convexity is a generalization of classical (linear) convexity to non-Euclidean (non-linear) geometry and metric spaces. Therefore, it is straightforward to show that all convex functions in Euclidean geometry are also g-convex, where the geodesics between pairs of matrices are simply line segments:
\begin{equation}
    {\zeta}_{l}({\bf p}_{0},{\bf p}_{1})= l{\bf p}_{0}+(1-l){\bf p}_{1} \label{eq:convexSet} \;.
\end{equation}
\end{rem} 
For the sake of brevity, we omit a
detailed theoretical introduction of g-convexity, and only borrow a result from \citep{zadeh2016geometric}. Interested readers are referred to \citep[Chapter 1]{wiesel2015structured} for a gentle introduction to this topic, and \citep[Chapter. 2]{papadopoulos2005metric}; \citep{rapcsak1991geodesic,ben1977generalized,liberti2004class,pallaschke2013foundations,bonnabel2009riemannian,moakher2005differential,sra2015conic,vishnoi2018geodesic} for more in-depth technical details.
Now we are ready to state the proof, which parallels the one provided in \citet[Theorem. 3]{zadeh2016geometric}.

\begin{proof}
We proceed in two steps. First, we consider P.D. manifolds and express \eqref{eq:g-convex-condition} in terms of geodesic paths and functions that lie on this particular space. We then show that $\mathcal{L}^{\mathrm{time}}_{\mathrm{conv}}({\bm \Gamma}^{k},{\bm \Lambda}^{k},{\bf B})$ is strictly g-convex on this specific domain. In the second step, we then derive the update rule proposed in \eqref{eq:BGeodesicUpdate}. 

\subsection{Part I: G-convexity of the majorizing cost function}
\label{subsec:g-convex-proof}
We consider geodesics along the P.D. manifold by setting $\zeta_l({\bf p}_{0},{\bf p}_{1})$ to $\xi_l({\bf B}_{0},{\bf B}_{1})$ as presented in Definition~\ref{def:PSD-geo-path}, and define $f(.)$ to be $f({\bf B}) = \mathrm{tr}\left(\left({\bf B}^{k}\right)^{-1}
{\bf B} \right) +\mathrm{tr}({\bf M}_{\mathrm{time}}^{k}{\bf B}^{-1})$, representing the cost function $\mathcal{L}^{\mathrm{time}}_{\mathrm{conv}}({\bm \Gamma}^{k},{\bm \Lambda}^{k},{\bf B})$.

We now show that $f({\bf B})$ is strictly g-convex on this specific domain. For continuous functions as considered in this paper, fulfilling \eqref{eq:g-convex-condition} for $f({\bf B})$ and $\xi_l({\bf B}_{0},{\bf B}_{1})$ with $l=\nicefrac{1}{2}$ is sufficient for strict g-convexity according to \emph{mid-point convexity} \citep{niculescu2006convex}:
\begin{align}
    &\mathrm{tr}\left(\left({\bf B}^{k}\right)^{-1}{\xi_{\nicefrac{1}{2}}({\bf B}_{0},{\bf B}_{1})}\right)+\mathrm{tr}\left({\bf M}_{\mathrm{time}}^{k}{\xi_{\nicefrac{1}{2}}({\bf B}_{0},{\bf B}_{1})}^{-1}\right) \nonumber \\ 
    &< \frac{1}{2}\mathrm{tr}\left(\left({\bf B}^{k}\right)^{-1}{{\bf B}_{0}}\right)+ \frac{1}{2}\mathrm{tr}\left({\bf M}_{\mathrm{time}}^{k}{{\bf B}_{0}}^{-1}\right) \nonumber \\ &\quad \quad + \frac{1}{2}\mathrm{tr}\left(\left({\bf B}^{k}\right)^{-1}{{\bf B}_{1}}\right)+ \frac{1}{2}\mathrm{tr}\left({\bf M}_{\mathrm{time}}^{k}{{\bf B}_{1}}^{-1}\right)\;.
    \label{eq:g-convex-function-trace}
\end{align}
Given $\left({\bf B}^{k}\right)^{-1} \in \mathcal{S}_{++}$, i.e., $\left({\bf B}^{k}\right)^{-1}>0$ and the operator inequality \citep[Chapter. 4]{bhatia2009positive}
\begin{align}
    \xi_{\nicefrac{1}{2}}({\bf B}_{0},{\bf B}_{1}) \prec \frac{1}{2}{\bf B}_{0}+ \frac{1}{2}{\bf B}_{1}\;,
    \label{eq:operator-ineq}
\end{align}
we have:
\begin{align}
    \mathrm{tr}\left(\left({\bf B}^{k}\right)^{-1}{\xi_{\nicefrac{1}{2}}({\bf B}_{0},{\bf B}_{1})}\right) &< \frac{1}{2}\mathrm{tr}\left(\left({\bf B}^{k}\right)^{-1}{{\bf B}_{0}}\right) + \frac{1}{2}\mathrm{tr}\left(\left({\bf B}^{k}\right)^{-1}{{\bf B}_{1}}\right) \;,
    \label{eq:proof-convex-main}
\end{align}
which is derived by multiplying both sides of Eq.~\eqref{eq:operator-ineq} with $\left({\bf B}^{k}\right)^{-1}$ followed by taking the trace on both sides. 

Similarly, we can write the operator inequality for $\{{\bf B}_{0}^{-1},{\bf B}_{1}^{-1}\}$ using Eq.~\eqref{eq:geodesic-PSD-inverse} as:
\begin{align}
    \xi_{\nicefrac{1}{2}}({\bf B}_{0},{\bf B}_{1})^{-1} =  \xi_{\nicefrac{1}{2}}({\bf B}_{0}^{-1},{\bf B}_{1}^{-1}) \prec \frac{1}{2}{\bf B}_{0}^{-1}+ \frac{1}{2}{\bf B}_{1}^{-1}\;.
    \label{eq:operator-ineq-inv}
\end{align}
Multiplying both sides of Eq.~\eqref{eq:operator-ineq-inv} by ${\bf M}_{\mathrm{time}}^{k} \in \mathcal{S}_{++}$ and applying the trace operator on both sides leads to:
\begin{align}
    \mathrm{tr}\left({\bf M}_{\mathrm{time}}^{k}{\xi_{\nicefrac{1}{2}}({\bf B}_{0},{\bf B}_{1})}^{-1}\right) &< \frac{1}{2}\mathrm{tr}\left({\bf M}_{\mathrm{time}}^{k}{{\bf B}_{0}}^{-1}\right)+ \frac{1}{2}\mathrm{tr}\left({\bf M}_{\mathrm{time}}^{k}{{\bf B}_{1}}^{-1}\right)\;. \label{eq:proof-convex-inverse}
\end{align}
Summing up \eqref{eq:proof-convex-main} and \eqref{eq:proof-convex-inverse} proves inequality~\eqref{eq:g-convex-function-trace} and concludes the first part of the proof.

\subsection{Part II: Derivation of the update rule in Eq.~\eqref{eq:BGeodesicUpdate}}
\label{subsection:update-rule-derivation}

We now present the second part of the proof by deriving the update rule in Eq.~\eqref{eq:BGeodesicUpdate}.
Since the cost function $\mathcal{L}^{\mathrm{time}}_{\mathrm{conv}}({\bm \Gamma}^{k},{\bm \Lambda}^{k},{\bf B})$ is strictly g-convex, its optimal solution in the $k$-th iteration is unique. More concretely, the optimum can be analytically derived by taking the derivative of Eq.~\eqref{eq:BGeodesicUpdate} and setting the result to zero as follows: 
\begin{align}
    \nabla \mathcal{L}^{\mathrm{time}}_{\mathrm{conv}}({\bm \Gamma}^{k},{\bm \Lambda}^{k},{\bf B}) &= \left({\bf B}^{k}\right)^{-1}-{\bf B}^{-1}{\bf M}_{\mathrm{time}}^{k}{\bf B}^{-1}=0 \;,
\end{align}
which results in 
\begin{align}
     {\bf B}\left({\bf B}^{k}\right)^{-1}{\bf B} &= {\bf M}_{\mathrm{time}}^{k}\;.
     \label{eq:Riccati}
\end{align}
This solution is known as the \emph{Riccati equation} and is the geometric mean between ${\bf B}^{k}$ and ${\bf M}_{\mathrm{time}}^{k}$ \citep{davis2007information,bonnabel2009riemannian}:
\begin{align*}
    {\bf B}^{k+1} \leftarrow ({\bf B}^{k})^{\frac{1}{2}}\left(({\bf B}^{k})^{\nicefrac{-1}{2}}{\bf M}_{\mathrm{time}}^{k}({\bf B}^{k})^{\nicefrac{-1}{2}}\right)^{\frac{1}{2}}({\bf B}^{k})^{\frac{1}{2}}\;.
\end{align*}
%
Deriving the update rule in Eq.~\eqref{eq:BGeodesicUpdate} concludes the second part of the proof of Theorem~\ref{Theo:Time-Sol}.
\end{proof}

\section{Proof of Theorem~\ref{Theo:Space-Surrogate}}
\label{appendix:Space-Surrogate}

\begin{proof}
Analogous to the proof of Theorem~\ref{Theo:Time-Surrogate} in Appendix~\ref{appendix:Time-Surrogate}, we start by recalling $\mathcal{L}_{\text{kron}}({\bm \Gamma},{\bm \Lambda},{\bf B})$ in Eq.~\eqref{eq:MLKronokCost}:
\begin{align*}
    \mathcal{L}_{\text{kron}}({\bm \Gamma},{\bm \Lambda},{\bf B}) &= T\log|{{\bm \Sigma}_{\bf y}}|+M\log|{\bf B}| +\frac{1}{G}\sum_{g=1}^{G}\text{tr}({{\bm \Sigma}_{\bf y}^{-1}}{\bf Y}_{g}{\bf B}^{-1}{\bf Y}_{g}^{\top}).
\end{align*}
Let ${\bf B }^{k}$ be the value of the temporal covariance matrix learned using Eq.~\eqref{eq:BGeodesicUpdate} at $k$-th iteration. Then, by ignoring the term $M\log|{\bf B}^{k}|$ that is only a function of ${\bf B}^{k}$, $\mathcal{L}_{\text{kron}}({\bm \Gamma},{\bm \Lambda},{\bf B})$ can be written as follows:
\begin{align}
    \mathcal{L}_{\text{kron}}^{\mathrm{space}}({\bm \Gamma},{\bm \Lambda},{\bf B }^{k}) 
    &= T\log|{{\bm \Sigma}_{\bf y}}| +\frac{1}{G}\sum_{g=1}^{G}\tr\left({{\bm \Sigma}_{\bf y}^{-1}}{\bf Y}_{g}({\bf B}^{k})^{-1}{\bf Y}_{g}^{\top}\right) \nonumber \\ 
    &= \log|{{\bm \Sigma}_{\bf y}}| +\frac{1}{TG}\sum_{g=1}^{G}\tr\left({{\bm \Sigma}_{\bf y}^{-1}}{\bf Y}_{g}({\bf B}^{k})^{-1}{\bf Y}_{g}^{\top}\right) \nonumber \\
    &= \log|{{\bm \Sigma}_{\bf y}}|+\tr\left({{\bm \Sigma}_{\bf y}^{-1}}\frac{1}{TG}\sum_{g=1}^{G}{\bf Y}_{g}({\bf B}^{k})^{-1}{\bf Y}_{g}^{\top}\right) \nonumber \\
    &= \log|{{\bm \Sigma}_{\bf y}}| +\tr\left({{\bm \Sigma}_{\bf y}^{-1}}{\bf M}_{\mathrm{space}}^{k}\right)\;,
    \label{eq:surrogate-derivation}
\end{align}
where ${\bf M}_{\mathrm{space}}^{k}:=\frac{1}{TG}\sum_{g=1}^{G}{\bf Y}_{g}({\bf B}^{k})^{-1}{\bf Y}_{g}^{\top}$. 

Similar to the argument made in Appendix \ref{appendix:Time-Surrogate}, a first order Taylor expansion of the log-determinant function around ${\bm \Sigma}_{\bf y}$ provides the following inequality:
\begin{align}
    \log\left|{\bm \Sigma}_{\bf y}\right| &\leq \log\left|{\bm \Sigma}_{\bf y}^{k}\right|+\tr\left(\left({\bm \Sigma}_{\bf y}^{k}\right)^{-1} \left({\bm \Sigma}_{\bf y}-{\bm \Sigma}_{\bf y}^{k}\right) \right) \nonumber \\ 
    &= \log\left|{\bm \Sigma}_{\bf y}^{k}\right| + \tr\left(\left({\bm \Sigma}_{\bf y}^{k}\right)^{-1}{\bm \Sigma}_{\bf y} \right)- \tr\left(\left({\bm \Sigma}_{\bf y}^{k}\right)^{-1}{\bm \Sigma}_{\bf y}^{k} \right) \nonumber \\ 
    &= \tr({\bm \Phi}^{\top}({\bm \Sigma}_{\bf y}^{k})^{-1}{\bm \Phi}{\bf H}) + \mathrm{const}
    \;,
    \label{eq:Taylor-MM-convex-bounding}
\end{align}
where the last step is derived using the augmented source and noise covariances, ${\bf H}:= [{\bm \Gamma}, \bf 0; \bf 0, {\bm \Lambda}]$, ${\bm \Phi}:=[{\bf L},{\bf I}]$ and ${\bm \Sigma}_{\bf y}={\bm \Phi}{\bf H}{\bm \Phi}^{\top}$. 

By inserting Eq.~\eqref{eq:surrogate-derivation} into Eq.~\eqref{eq:Taylor-MM-convex-bounding}, the first term of Eq.~\eqref{eq:SpaceSurrogateFunction}, $\tr({\bm \Phi}^{\top}({\bm \Sigma}_{\bf y}^{k})^{-1}{\bm \Phi}{\bf H})$, can be directly inferred:
\begin{align}
    \mathcal{L}_{\text{kron}}^{\mathrm{space}}({\bm \Gamma},{\bm \Lambda},{\bf B }^{k}) = \mathcal{L}_{\text{kron}}^{\mathrm{space}}({\bf H},{\bf B }^{k}) 
    &= \log|{{\bm \Sigma}_{\bf y}}| +\tr\left({{\bm \Sigma}_{\bf y}^{-1}}{\bf M}_{\mathrm{space}}^{k}\right)  \nonumber \\
    &\leq \tr({\bm \Phi}^{\top}({\bm \Sigma}_{\bf y}^{k})^{-1}{\bm \Phi}{\bf H}) +\tr\left({{\bm \Sigma}_{\bf y}^{-1}}{\bf M}_{\mathrm{space}}^{k}\right) + \text{const}\;,
    \label{eq:Space-Surrogate-derivation}
\end{align}
We further show how the second term in Eq.~\eqref{eq:SpaceSurrogateFunction} can be derived. To this end, we construct an upper bound on $\tr\left({{\bm \Sigma}_{\bf y}^{-1}}{\bf M}_{\mathrm{space}}^{k}\right)$ using an inequality derived from the Schur complement of ${\bm \Sigma}_{\bf y}$. Before presenting this inequality, the subsequent definition of the Schur complement of matrix ${\bm \Sigma}_{\bf y}$ is required: 

\begin{defi} 
\label{def:propSchur-H} 
For a positive semidefinite (PSD) matrix ${\bm \Sigma}_{\bf y}$, and a partitioning 
\begin{align}
    {\bf X} &=\left[\begin{array}{cc}
    {\bf D} & {\bf G}\\
    {\bf G}^{\top} & {\bf B}
    \end{array}\right] \;,
    \label{eq:XPartition-H}
\end{align}
its Schur complement is defined as
\begin{align}
    {\bf S} &:={\bf D}-{\bf G}{{\bm \Sigma}_{\bf y}^{-1}}{\bf G}^{\top}\label{eq:SchurComplement-H} \\
\end{align}
The Schur complement condition states that the matrix ${\bf X}$ is PSD, ${\bf X}\geq {\bf 0}$, if and only if the
Schur complement of ${\bm \Sigma}_{\bf y}$ is PSD, ${\bf S}\geq {\bf 0}$.
\end{defi}

Now we are ready to construct an upper bound on $\tr\left({{\bm \Sigma}_{\bf y}^{-1}}{\bf M}_{\mathrm{space}}^{k}\right)$. To this end, we show that $\tr\left({{\bm \Sigma}_{\bf y}^{-1}}{\bf M}_{\mathrm{space}}^{k}\right)$ can be majorized as follows:
\begin{align}
    \tr\left({{\bm \Sigma}_{\bf y}^{-1}}{\bf M}_{\mathrm{space}}^{k}\right) &\leq \tr({\bf H}^{k}{\bm \Phi}^{\top}({\bm \Sigma}_{\bf y}^{k})^{-1}{\bf M}_{\mathrm{space}}^{k}({\bm \Sigma}_{\bf y}^{k})^{-1}{\bm \Phi}{\bf H}^{k}{\bf H}^{-1})  \;.
    \label{eq:secondterm-H}
\end{align}

By defining $\mathbf{V}$ as: 
\begin{align}
    \mathbf{V} &=\left[\begin{array}{c}
    ({\bm \Sigma}_{\bf y}^{k})^{-1}{\bm \Phi}{\bf H}^{k}{\bf H}^{\frac{-1}{2}}\\
    {\bm \Phi}{\bf H}^{\frac{1}{2}}
    \end{array}\right]\;, \label{eq:Vdef-H}
\end{align}
the PSD property of $\mathbf{S}$ can be inferred as:
\begin{align}
    \mathbf{S} &= \left[\begin{array}{cc}
    ({\bm \Sigma}_{\bf y}^{k})^{-1}{\bm \Phi}{\bf H}^{k}{\bf H}^{-1}{\bf H}^{k}{\bm \Phi}^{\top}({\bm \Sigma}_{\bf y}^{k})^{-1} & \textbf{I} \\
    \textbf{I} & {\bm \Phi}{\bf H}{\bm \Phi}^{\top}
    \end{array}\right]=\mathbf{V}\mathbf{V}^{\top}\geq0\;. \label{eq:Sdef-H}
\end{align}
By employing the definition of the Schur complement with ${\bf D}=({\bm \Sigma}_{\bf y}^{k})^{-1}{\bm \Phi}{\bf H}^{k}{\bf H}^{-1}{\bf H}^{k}{\bm \Phi}^{\top}({\bm \Sigma}_{\bf y}^{k})^{-1}$, ${\bf G} = \textbf{I}$ and ${\bm \Sigma}_{\bf y}={\bm \Phi}{\bf H}{\bm \Phi}^{\top}$, we have:
\begin{equation}
     ({\bm \Sigma}_{\bf y}^{k})^{-1}{\bm \Phi}{\bf H}^{k}{\bf H}^{-1}{\bf H}^{k}{\bm \Phi}^{\top}({\bm \Sigma}_{\bf y}^{k})^{-1} \geq \left({\bm \Phi}{\bf H}{\bm \Phi}^{\top} \right)^{-1} \;.
     \label{eq:afterSchur-H}
\end{equation}
The inequality in Eq.~\eqref{eq:secondterm-H} can be directly inferred by multiplying ${\bf M}_{\mathrm{space}}^{k}$ to both sides of Eq.~\eqref{eq:afterSchur-H}, applying trace operator, and rearranging the arguments in the trace operator:
\begin{align}
    \tr({\bf M}_{\mathrm{space}}^{k}{\bm \Sigma}_{\bf y}^{-1}] &\leq \tr({\bf M}_{\mathrm{space}}^{k}({\bm \Sigma}_{\bf y}^{k})^{-1}{\bm \Phi}{\bf H}^{k}{\bf H}^{-1}{\bf H}^{k}{\bm \Phi}^{\top}({\bm \Sigma}_{\bf y}^{k})^{-1})
    \nonumber \\  
    &= \tr({\bf H}^{k}{\bm \Phi}^{\top}({\bm \Sigma}_{\bf y}^{k})^{-1}{\bf M}_{\mathrm{space}}^{k}({\bm \Sigma}_{\bf y}^{k})^{-1}{\bm \Phi} {\bf H}^{k} {\bf H}^{-1})
    \nonumber \\  
    &= \tr\left({\bf M}_{\mathrm{SN}}^{k} {\bf H}^{-1} \right)\;.  \label{eq:afterSchurfinal-H}    
\end{align}
%

By inserting Eq.~\eqref{eq:afterSchurfinal-H} into Eq.~\eqref{eq:Space-Surrogate-derivation}, we have
\begin{align}
    \mathcal{L}_{\text{kron}}^{\mathrm{space}}({\bm \Gamma},{\bm \Lambda},{\bf B }^{k}) = \mathcal{L}_{\text{kron}}^{\mathrm{space}}({\bf H},{\bf B }^{k}) 
    &\leq \tr({\bm \Phi}^{\top}({\bm \Sigma}_{\bf y}^{k})^{-1}{\bm \Phi}{\bf H}) +\tr\left({{\bm \Sigma}_{\bf y}^{-1}}{\bf M}_{\mathrm{space}}^{k}\right) + \text{const} \nonumber \\ 
    &\leq \tr({\bm \Phi}^{\top}({\bm \Sigma}_{\bf y}^{k})^{-1}{\bm \Phi}{\bf H}) +  \tr\left({\bf M}_{\mathrm{SN}}^{k} {\bf H}^{-1} \right) + \text{const} \nonumber \\ 
    &= \mathcal{L}^{\mathrm{space}}_{\mathrm{conv}}({\bf H},{\bf B}^{k}) + \text{const} \;.
    \label{eq:Space-Surrogate-derivation-final}
\end{align}
Note that constant values in \eqref{eq:Space-Surrogate-derivation-final} do not depend on ${\bf H}$; hence, they can be ignored in the optimization procedure. We have shown that minimizing Eq.~\eqref{eq:MLKronokCost} with respect to ${\bf H}$ is equivalent to minimizing $\mathcal{L}^{\mathrm{space}}_{\mathrm{conv}}({\bf H},{\bf B}^{k})$, which concludes the proof.
\end{proof}

\section{Proof of Theorem~\ref{Theo:Space-Sol}}
\label{appendix:Space-Solution}

\begin{proof}
{We proceed in two steps. First, we show that $\mathcal{L}^{\mathrm{space}}_{\mathrm{conv}}({\bf H},{\bf B}^{k})$ is convex in ${\bf h}$. Then, we derive the update rule proposed in Eq.~\eqref{eq:HDiagonalUpdate}.}

\subsection{Part I: Convexity of the majorizing cost function}
We start the proof by constraining ${\bf  H}$ to the set of diagonal matrices with non-negative entries $\mathcal{S}$, i.e., $\mathcal{S}=\{{\bf H}\; | \; {\bf H} =\mathrm{diag}({\bf  h})=\mathrm{diag}([h_{1},\dots,h_{N+M}]^\top),\;h_{n} \geq 0,\;\text{for}\;i=1,\hdots,N+M\}$. We continue by reformulating the constrained optimization with respect to the source covariance matrix,
\begin{align}
    {\bf H}^{k+1} &=  \argmin_{{\bf H} \in \mathcal{S},\;{\bf B}= {\bf B}^{k} } \mathrm{tr}\left({\bm \Phi}^{\top}\left({\bm \Sigma}_{\bf y}^{k}\right)^{-1}{\bm \Phi} {\bf  H}\right)+\mathrm{tr}({\bf M}_{\mathrm{SN}}^{k}{\bf H}^{-1})\;,
    \label{eq:Gamma-Constrained-Proof}
\end{align}
as follows: 
\begin{align}
    {\bf h}^{k+1} &=  \argmin_{{\bf  h} \geq 0,\;{\bf B}= {\bf B}^{k}} \quad \underbrace{\mathrm{diag}\left({\bm \Phi}^{\top}\left({\bm \Sigma}_{\bf y}^{k}\right)^{-1}{\bm \Phi} \right){\bf  h}+\mathrm{diag}\left({\bf M}_{\mathrm{SN}}^{k}\right) {\bf  h}^{-1}}_{\mathcal{L}^{\mathrm{space}}_{\mathrm{diag}}({\bf  h}|{\bf  h}^{k})}\;,
    \label{eq:SN-diag}
\end{align}
where ${\bf  h}^{-1}=[h_{1}^{-1},\dots,h_{N}^{-1}]^\top$ is defined as the element-wise inversion of ${\bf h}$. {Let ${\bf V}^k:={\bm \Phi}^{\top} \left({\bm \Sigma}_{\bf y}^{k}\right)^{-1}{\bm \Phi}$. Then, we rewrite $\mathcal{L}^{\mathrm{space}}_{\mathrm{diag}}({\bf h}|{\bf h}^{k})$ as   
\begin{align}
    \mathcal{L}^{\mathrm{space}}_{\mathrm{diag}}({\bf h}|{\bf h}^{k}) &=  \mathrm{diag}\left({\bf V}^k \right){\bf  h}+\mathrm{diag}\left({\bf M}_{\mathrm{SN}}^{k}\right) {\bf  h}^{-1}\;.
    \label{eq:SN-diag-appendix}
\end{align}
The convexity of  $\mathcal{L}^{\mathrm{space}}_{\mathrm{diag}}({\bf h}|{\bf h}^{k})$ can be directly inferred from the convexity of $\mathrm{diag}\left[{\bf V}^k \right]{\bf h}$ and $\mathrm{diag}\left[{\bf M}_{\mathrm{SN}}^{k}\right]{\bf h}^{-1}$ with respect to ${\bf h}$ \citep[Chapter. 3]{boyd2004convex}.}
\subsection{Part II: Derivation of the update rule in Eq.~\eqref{eq:HDiagonalUpdate}}
\label{appendix:H-derivation}
We now present the second part of the proof by deriving the update rule in Eq.~\eqref{eq:HDiagonalUpdate}.
Since the cost function $\mathcal{L}^{\mathrm{space}}_{\mathrm{diag}}({\bf h}|{\bf h}^{k})$ is convex, its optimal solution in the $k$-th iteration is unique. Therefore, the optimization with respect to heteroscedastic source and noise variances is carried out by taking the derivative of \eqref{eq:SN-diag} with respect to $h_{i},\;\text{for}\; n=1,\hdots,M+N$, and setting it to zero:
\begin{align*}
    \frac{\partial}{\partial h_{i}} & \left(   
    \left[{\bm \Phi}^{\top}\left({\bm \Sigma}_{\bf y}^{k}\right)^{-1}{\bm \Phi} \right]{h}_i + \left[{\bf M}_{\mathrm{SN}}^{k} \right]{h}_{i}^{-1} \right)  \\
    &=\left[{\bm \Phi}^{\top}\left({\bm \Sigma}_{\bf y}^{k}\right)^{-1}{\bm \Phi} \right]_{i,i} - \frac{1}{(h_{i})^2} \left[{\bf M}_{\mathrm{SN}}^{k} \right]_{i,i}  \\ &= 0 \quad \text{for}\;i=1,\hdots,N+M\;,
\end{align*}
where ${\bm \Phi}_{i}$ denotes the $n$-th column of the augmented lead field matrix. This yields the following update rule:
\begin{align}
    {\bf H}^{k+1} =\mathrm{diag}({\bf  h}^{k+1}),\; \text{where}, \quad h_{i}^{k+1} &\leftarrow \sqrt{\frac{\left[{\bf M}_{\mathrm{SN}}^{k} \right]_{i,i}}{\left[{\bm \Phi}^{\top}\left({\bm \Sigma}_{\bf y}^{k}\right)^{-1}{\bm \Phi} \right]_{i,i}}} = \sqrt{\frac{ \frac{1}{T} \sum_{t=1}^{T} (\bar{\bm \eta}^{k}_{n}(t))^{2}}{{\bm \Phi}_{n}^{\top}\left({\bm \Sigma}_{\bf y}^{k}\right)^{-1}{\bm \Phi}_{i}}} \nonumber \\ &\quad \text{for}\;i=1,\hdots,N+M \;. 
    \label{eq:H-derivation}
\end{align} 
The updates rule in Eq.~\eqref{eq:HDiagonalUpdate} can be directly inferred by defining ${\bf g} :=\mathrm{diag}({\bf M}_{\mathrm{SN}}^{k})$ and ${\bf z} :=\mathrm{diag}({\bm \Phi}^{\top}({\bm \Sigma}_{\bf y}^{k})^{-1}{\bm \Phi})$, which leads to: $g_{i}^{k} = \left[{\bf M}_{\mathrm{SN}}^{k} \right]_{i,i}$ and $z_{i}^{k}=\left[{\bm \Phi}^{\top}\left({\bm \Sigma}_{\bf y}^{k}\right)^{-1}{\bm \Phi} \right]_{i,i}$. This concludes the proof. 
\end{proof}



\section{Champagne with heteroscedastic noise learning}
\label{appendix:Champ-Hetero-deriv}
Interestingly, identical update rules as those proposed in Champagne \citep{wipf2010robust} and heteroscedastic noise learning \citep{cai2021robust} can be derived for source and noise variances, respectively, by selecting the corresponding indices of matrix ${\bf H}$ associated to noise and source covariances. 
\subsection{Update rule for source variances}
Given $\left[{\bm \Phi} \right]_{1:M,1:N}={\bf L}$, $\left[{\bf H} \right]_{1:N,1:N} = {\bm \Gamma}$, and $\left[\bar{\bm \eta}(t)\right]_{1:N} = \bar{\bf x}(t)$, the update rule for ${\bm \Gamma}^{k+1} =\mathrm{diag}({\bm \gamma}^{k+1})$ is derived by replacing ${\bf H}$, ${\bm \Phi}$ and $\bar{\bm \eta}^{k}_{n}(t)$ in Eq.~\eqref{eq:H-derivation} with ${\bm \Gamma}$, ${\bf L}$ and $\bar{\bf x}^{k}_{n}(t)$, respectively, and defining the counterpart of ${\bf M}_{\mathrm{SN}}^{k}$ for sources accordingly as ${\bf M}_{\mathrm{S}}^{k}:= {\bm \omega}_{\mathrm{S}}^{k}{\bf M}_{\mathrm{space}}^{k}({\bm \omega}_{\mathrm{S}}^{k})^\top$, where ${\bm \omega}_{\mathrm{S}}^{k}:={\bm \Gamma}^{k}{\bf L}^{\top}({\bm \Sigma}_{\bf y}^{k})^{-1}$. The update rule for the source variances is then obtained as follows:
\begin{align}
    \gamma_{n}^{k+1} \leftarrow \sqrt{\frac{\left[{\bf M}_{\mathrm{S}}^{k} \right]_{n,n}}{\left[{\bf L}^{\top}\left({\bm \Sigma}_{\bf y}^{k}\right)^{-1}{\bf L} \right]_{n,n}}} &= \sqrt{\frac{\frac{1}{T} \sum_{t=1}^{T} (\bar{\bf x}^{k}_{n}(t))^{2}}{{\bf L}_{n}^{\top}\left({\bm \Sigma}_{\bf y}^{k}\right)^{-1}{\bf L}_{n}}} \nonumber \\ &\quad  \text{for}\;n=1,\hdots,N \;, 
    \label{eq:gamma-diag-update}
\end{align} 
where ${\bf L}_{n}$ denotes the $n$-th column of the lead field matrix.

\subsection{Update rule for noise variances}

Similarly, given $\left[{\bm \Phi} \right]_{1:M,N+1:N+M}={\bf I}$, $\left[{\bf H} \right]_{N+1:N+M,N+1:N+M} = {\bm\Lambda}$, and $\left[\bar{\bm \eta}(t)\right]_{N+1:N+M} = \bar{\bf e}(t):={\bf y}(t)-{\bf L}\bar{\bf x}(t)$, the update rule for ${\bm \Lambda}^{k+1} =\mathrm{diag}({\bm \lambda}^{k+1})$ is derived by replacing ${\bf H}$, ${\bm \Phi}$ and $\bar{\bm \eta}^{k}_{n}(t)$ in Eq.~\eqref{eq:H-derivation} with ${\bm \Lambda}$, ${\bf I}$ and $\bar{\bf e}^{k}_{n}(t)$, respectively, and defining the counterpart of ${\bf M}_{\mathrm{SN}}^{k}$ for the noise accordingly as ${\bf M}_{\mathrm{N}}^{k}:= {\bm \omega}_{\mathrm{N}}^{k}{\bf M}_{\mathrm{space}}^{k}({\bm \omega}_{\mathrm{N}}^{k})^\top$ with ${\bm \omega}_{\mathrm{N}}^{k} = {\bm \Lambda}^{k}({\bm \Sigma}_{\bf y}^{k})^{-1}$. The update rule for the noise variances is then derived as follows:
\begin{align}
    \lambda_{m}^{k+1} \leftarrow \sqrt{\frac{\left[{\bf M}_{\mathrm{N}}^{k} \right]_{m,m}}{\left[\left({\bm \Sigma}_{\bf y}^{k}\right)^{-1} \right]_{m,m}}} &= \sqrt{\frac{\sum_{t=1}^{T} (\bar{\bf e}^{k}_{n}(t))^{2}}{\left[\left({\bm \Sigma}_{\bf y}^{k}\right)^{-1}\right]_{m,m}}} \nonumber \\ &\quad \text{for}\;m=1,\hdots,M \;,
\end{align}
which is identical to the update rule of the Champagne with heteroscedastic noise learning as presented in \citet{cai2021robust}. 

{
\section{Proof of Theorem~\ref{theo:MM-Convergence-Guarantees}}
\label{appendix:MM-Convergence-Guarantees}
We prove Theorem \ref{theo:MM-Convergence-Guarantees} by showing that the alternating update rules for ${\bf  B}$ and ${\bf  H}$, Eqs.~\eqref{eq:BGeodesicUpdate} and \eqref{eq:HDiagonalUpdate}, are guaranteed to converge to a stationary point of the Bayesian Type-II likelihood $\mathcal{L}_{\text{kron}}({\bm \Gamma},{\bm \Lambda},{\bf B})$ Eq.~\eqref{eq:MLKronokCost}. More generally, we prove that full Dugh is an instance of the general class of majorization-minimization (MM) algorithms, for which this property follows by construction. To this end, we first briefly review theoretical concepts behind the majorization-minimization (MM) algorithmic framework \citep{hunter2004tutorial,razaviyayn2013unified,jacobson2007expanded,wu2010mm}.
} 

\subsection{Required conditions for majorization-minimization algorithms}
MM is a versatile framework for optimizing general non-linear optimization programs. The main idea behind MM is to replace the original cost function in each iteration by an upper bound, also known as majorizing function, whose minimum is easy to find. Compared to other popular optimization paradigms such as (quasi)-Newton methods, MM algorithms enjoy guaranteed convergence to a stationary point \citep{sun2017majorization}. The MM class covers a broad range of common optimization algorithms such as \emph{convex-concave procedures (CCCP)} and \emph{proximal methods}   \citep[Section~IV]{sun2017majorization}, \citep{mjolsness1990algebraic,yuille2003concave,lipp2016variations}. Such algorithms have been applied in various domains such as non-negative matrix factorization \citep{Fagot2019}, graph learning \citep{kumar2020unified}, robust portfolio optimization in finance \citep{feng2016signal}, direction of arrival (DoA) and channel estimation in wireless communications \citep{prasad2015joint,gerstoft2016multisnapshot,haghighatshoar2017massive,khalilsarai2020structured}, internet of things (IoT) \citep{fengler2019massive,fengler2019non}, and brain source imaging \citep{luessi2013sparse,hashemi2018improving,bekhti2018hierarchical,cai2021robust,hashemi2021unification}. Interested readers are referred to \citet{sun2017majorization} for an extensive list of applications on MM.

We define an original optimization problem with the objective of minimizing a continuous function $f({\bf u})$ within a closed convex set $\mathcal{U} \subset \mathbb{R}^n $:
\begin{equation}
    \label{eq:OriginalNonconvex}
    \begin{aligned}
        & \underset{{\bf u}}{\text{min}} \; f({\bf u}) \quad \text{subject to } {\bf u} \in \mathcal{U} \;.
    \end{aligned}
\end{equation}
Then, the idea of MM can be summarized as follows. First, construct a continuous \emph{surrogate function} $g({\bf u}|{\bf u}^{k})$ that \emph{majorizes}, or upper-bounds, the original function $f({\bf u})$ and coincides with $f({\bf u})$ at a given point ${\bf u}^{k}$:
\begin{align*}
    &\text{[A1]} & &g({\bf u}^{k}|{\bf u}^{k})=f({\bf u}^{k}) & &\forall \; {\bf u}^{k}\in\mathcal{U} \\
    &\text{[A2]} & &g({\bf u}|{\bf u}^{k}) \geq f({\bf u}) &  &\forall \; {\bf u},{\bf u}^{k}\in\mathcal{U}\;. 
\end{align*}
Second, starting from an initial value ${\bf u}^{0}$, generate a sequence of feasible points ${\bf u}^{1}, {\bf u}^{2}, \hdots, {\bf u}^{k}, {\bf u}^{k+1}$ as solutions of a series of successive simple optimization problems, where
\begin{align*}
    &\text{[A3]} &{\bf u}^{k+1} := \text{arg} \underset{{\bf u} \in \mathcal{U} }{\text{ min}}\;g({\bf u}|{\bf u}^{k})\;.
\end{align*}
\begin{defi}
    \label{Def:MM}
Any algorithm fulfilling conditions [A1]--[A3] is called a Majorization Minimization (MM) algorithm.
\end{defi}
If a surrogate function fulfills conditions [A1]--[A3], then the value of the cost function $f$ decreases in each iteration:
\begin{cor}
    \label{cor:descending}
    An MM algorithm has a \emph{descending trend} property, whereby the value of the cost function $f$ decreases in each iteration: $f({\bf u}^{k+1}) \leq f({\bf u}^{k})$.
\end{cor}
\begin{proof}
To verify the descending trend in the MM framework, it is sufficient to show that $f({\bf u}^{k+1}) \leq f({\bf u}^{k})$. To this end, we have $f({\bf u}^{k+1}) \leq g({\bf u}^{k+1}|{\bf u}^{k} )$ from condition [A2]. Condition [A3] further states that $g({\bf u}^{k+1}|{\bf u}^{k})\leq g({\bf u}^{k}|{\bf u}^{k})$, while $g({\bf u}^{k}|{\bf u}^{k} )=f({\bf u}^{k})$ holds according to [A1]. Putting everything together, we have:
\begin{align*}
    f({\bf u}^{k+1}) \stackrel{\text{[A2]}}{\leq} g({\bf u}^{k+1}|{\bf u}^{k}) \stackrel{\text{[A3]}}{\leq} g({\bf u}^{k}|{\bf u}^{k} ) \stackrel{\text{[A1]}}{=} f({\bf u}^{k})\;,
\end{align*}
which concludes the proof. 
\end{proof}
While Corollary~\ref{cor:descending} guarantees a descending trend, convergence requires additional assumptions on particular properties of $f$ and $g$ \citep{jacobson2007expanded,razaviyayn2013unified}. For the smooth functions considered in this paper, we require that the derivatives of the original and surrogate functions coincide at ${\bf u}^{k}$:
\begin{align*}
     &\text{[A4]} & &\nabla g({\bf u}^{k}|{\bf u}^{k})= \nabla f({\bf u}^{k}) & &\forall \; {\bf u}^{k}\in\mathcal{U} \;.
\end{align*}
We can then formulate the following, stronger, theorem:
\begin{theo}
    \label{Theo:stationary}
    For an MM algorithm that additionally satisfies [A4], every limit point of the sequence of minimizers generated through [A3] is a stationary point of the original optimization problem Eq.~\eqref{eq:OriginalNonconvex}.  
\end{theo}
\begin{proof}
A detailed proof is provided in \citet[Theorem 1]{razaviyayn2013unified}.
\end{proof}
Note that since we are working with smooth functions, conditions [A1]--[A4] are sufficient to prove convergence to a stationary point according to Theorem~\ref{Theo:stationary}.

\subsection{Detailed derivation of the proof of Theorem \ref{theo:MM-Convergence-Guarantees}} 
We now show that full Dugh is an instance of majorization-minimization as defined above, which fulfills Theorem~\ref{Theo:stationary}.
\begin{proof}
We need to prove that conditions [A1]--[A4] are fulfilled for full Dugh. To this end, we first prove conditions [A1]--[A4] for the optimization with respect to ${\bf B}$ based on the convex surrogate function in Eq.~\eqref{eq:TimeSurrogateFunction}, $\mathcal{L}^{\mathrm{time}}_{\mathrm{conv}}({\bm \Gamma}^{k},{\bm \Lambda}^{k},{\bf B})$. For this purpose, we recall the upper bound on $\log|{\bf B}|$ in Eq.~\eqref{eq:Time-Surrogate-derivation}, which fulfills condition [A2] since it majorizes $\log |{\bf B}|$ as a result of the concavity of the log-determinant function and its first-order Taylor expansion around ${\bf B}^{k}$. Besides, it automatically satisfies conditions [A1] and [A4] by construction, because the majorizing function in Eq.~\eqref{eq:Time-Surrogate-derivation} is obtained through a Taylor expansion around ${\bf B}^{k}$. Concretely, $\text{[A1]}$ is satisfied because the equality in Eq.~\eqref{eq:Time-Surrogate-derivation} holds for ${\bf B}={\bf B}^{k}$. Similarly, $\text{[A4]}$ is satisfied because the gradient of $\log\left|{\bf B}\right|$ at point ${\bf B}^{k}$, $\left({\bf B}^{k}\right)^{-1}$ defines the linear Taylor approximation $\log\left|{\bf B}^{k}\right| + \mathrm{tr}\left(\left({\bf B}^{k}\right)^{-1} \left({\bf B} - {\bf B}^{k}\right) \right)$. Thus, both gradients coincide in ${\bf B}^{k}$ by construction. We can further prove that [A3] can be satisfied by showing that $\mathcal{L}^{\mathrm{time}}_{\mathrm{conv}}({\bm \Gamma}^{k},{\bm \Lambda}^{k},{\bf B})$ reaches its global minimum in each MM iteration. This is guaranteed if $\mathcal{L}^{\mathrm{time}}_{\mathrm{conv}}({\bm \Gamma}^{k},{\bm \Lambda}^{k},{\bf B})$ can be shown to be convex or g-convex with respect to ${\bf B}$. To this end, we first require the subsequent proposition:
\begin{prop}
    \label{prop:gconvex-global}
    Any local minimum of a g-convex function over a g-convex set is a global minimum.
\end{prop}
\begin{proof}
    A detailed proof is presented in \citet[Theorem 2.1]{rapcsak1991geodesic}.
\end{proof}
Given the proof presented in Appendix~\ref{subsec:g-convex-proof}, we can conclude that $\mathcal{L}^{\mathrm{time}}_{\mathrm{conv}}({\bf H}^{k},{\bf B})$ is g-convex; hence, any local minimum of $\mathcal{L}^{\mathrm{time}}_{\mathrm{conv}}({\bf H}^{k},{\bf B})$ is a global minimum according to Proposition~\ref{prop:gconvex-global}. This proves that condition [A3] is fulfilled and completes the proof that the optimization of Eq.~\eqref{eq:MLKronokCost} with respect to ${\bf B}$ using the convex surrogate cost function Eq.~\eqref{eq:TimeSurrogateFunction} leads to an MM algorithm.

The proof of conditions [A1], [A2] and [A4] for the optimization with respect to ${\bf H}$ based on the convex surrogate function in Eq.~\eqref{eq:SpaceSurrogateFunction}, $\mathcal{L}^{\mathrm{space}}_{\mathrm{conv}}({\bf H},{\bf B}^{k})$, can be presented analogously. To this end, we recall the upper bound on $\log\left|{\bm \Sigma}_{\bf y}\right|$ in Eq.~\eqref{eq:Taylor-MM-convex-bounding}, which fulfills condition [A2] since it majorizes $\log\left|{\bm \Sigma}_{\bf y}\right|$ as a result of the concavity of the log-determinant function and its first-order Taylor expansion around ${\bm \Sigma}_{\bf y}^{k}$. Besides, it automatically satisfies conditions [A1] and [A4] by construction, because the majorizing function in Eq.~\eqref{eq:Taylor-MM-convex-bounding} is obtained through a Taylor expansion around ${\bm \Sigma}_{\bf y}^{k}$. Concretely, $\text{[A1]}$ is satisfied because the equality in Eq.~\eqref{eq:Taylor-MM-convex-bounding} holds for ${\bm \Sigma}_{\bf y}={\bm \Sigma}_{\bf y}^{k}$. Similarly, $\text{[A4]}$ is satisfied because the gradient of $\log\left|{\bm \Sigma}_{\bf y}\right|$ at point ${\bm \Sigma}_{\bf y}^{k}$, $\left({\bm \Sigma}_{\bf y}^{k}\right)^{-1}$ defines the linear Taylor approximation $\log\left|{\bm \Sigma}_{\bf y}^{k}\right| + \mathrm{tr}\left[\left({\bm \Sigma}_{\bf y}^{k}\right)^{-1} \left({\bm \Sigma}_{\bf y} - {\bm \Sigma}_{\bf y}^{k}\right) \right]$. Thus, both gradients coincide in ${\bm \Sigma}_{\bf y}^{k}$ by construction. We can further prove that [A3] can be satisfied by showing that $\mathcal{L}^{\mathrm{time}}_{\mathrm{conv}}({\bf H}^{k},{\bf B})$ reaches its global minimum in each MM iteration. This is guaranteed if $\mathcal{L}^{\mathrm{time}}_{\mathrm{conv}}({\bf H}^{k},{\bf B})$ can be shown to be convex with respect to ${\bf H} =\mathrm{diag}({\bf h})$. Given the proof presented in Appendix~\ref{appendix:H-derivation}, we can show that [A3] is also satisfied since $\mathcal{L}^{\mathrm{space}}_{\mathrm{conv}}({\bf H},{\bf B}^{k})$ in Eq.~\eqref{eq:SpaceSurrogateFunction} is a convex function with respect to ${\bf h}$. The convexity of $\mathcal{L}^{\mathrm{space}}_{\mathrm{conv}}({\bf H},{\bf B}^{k})$, which ensures that condition [A3] can be satisfied using standard optimization, along with the fulfillment of conditions [A1], [A2] and [A4], ensure that Theorem~\ref{Theo:stationary} holds for $\mathcal{L}^{\mathrm{space}}_{\mathrm{conv}}({\bf H},{\bf B}^{k})$. This completes the proof that the optimization of Eq.~\eqref{eq:MLKronokCost} with respect to ${\bf H}$ using the convex surrogate cost function Eq.~\eqref{eq:SpaceSurrogateFunction} leads to an MM algorithm that is guaranteed to converge.
\end{proof}

\section{Proof of Proposition~\ref{prop:toeplitz}}
\label{appendix:toeplitz}

This is a well-established classical results of the signal processing literature. Therefore, we only provide two remarks highlighting important connections between Preposition~\ref{prop:toeplitz} and our proposed method, and refer the interested reader to \citet[Chapter 5]{grenander1958toeplitz} for a detailed proof of the theorem.
\begin{rem}
    As indicated in \citet{babu2016melt}, the set of embedded circulant matrices with size $L \times L$, $\mathcal{B}^L$, does not constist exclusively of positive definite Toeplitz matrices. Therefore, we restrict ourselves to embedded circulant matrices ${\bf P}$ that are also positive definite. This restriction indeed makes the diagonalization inaccurate, but we can improve the accuracy by choosing a large $L$. Practical evaluations have shown that choosing $L \geq 2T-1$ provides sufficient approximation result.
\end{rem}

\begin{rem}
The $\emph{Carath\'eodory}$ $\emph{parametrization}$ of a Toeplitz matrix \citep[Section 4.9.2]{stoica2005spectral} states that any PD matrix can be represented as ${\bf B}={\bf A}{\bf P}^{'}{\bf A}^{H}$ with $[{\bf A}]_{m,l}=e^{i(w_{l})(m-1)}$ and ${\bf P}^{'}=\mathrm{diag}(p_{0}^{'},p_{1}^{'},\dots,p_{L-1}^{'})$ where $w_{l}$ and $p_{l}$ are specific frequencies and their corresponding amplitudes. By comparing the $\emph{Carath\'eodory}$ $\emph{parametrization}$ of ${\bf B}$ with its Fourier diagonalization (Eq.~\eqref{eq:DFT}), it can be seen that Fourier diagonalization force the frequencies to lie on the Fourier grid, i.e. $w_{l} = \frac{2\pi(l-1)}{L}$ , which indeed makes the diagonalization slightly inaccurate. The approximation accuracy can, however, be improved by increasing $L$. The Szeg{\"o} theorem \citep{grenander1958toeplitz,gray2006toeplitz} states that a Toeplitz matrix is asymptotically ($L \to \infty$) diagonalized by the DFT matrix.
\end{rem}

\section{Proof of Theorem \ref{Theo:Time-Toeplitz}}
\label{appendix:Time-Toeplitz}
\begin{proof} 
We proceed in two steps. First, we show that the cost function in Eq.~\eqref{eq:MinComstrainedTempo} is convex with respect to ${\bf p}$. In the second step, we then derive the update rule proposed in \eqref{eq:BToeplitzUpdate}.
\subsection{Part I: Convexity of the majorizing cost function}
The proof of this section parallels the one provided in \citep[Proposition 4]{sun2016robust}.
We start by recalling Eq.~\eqref{eq:MinComstrainedTempo}:
\begin{align}
    {\bf B}^* &=\argmin_{{\bf B} \in \mathcal{B},\;{\bf H}= {\bf H}^{k} } \tr(({\bf B}^{k})^{-1}{\bf B})+\tr({\bf M}_{\mathrm{time}}^{k}{\bf B}^{-1})\;.
    \label{eq:MinComstrainedTempo-appendix}
\end{align}
We then show that the second term in Eq.~\eqref{eq:MinComstrainedTempo-appendix} can be upper-bounded as follows:
\begin{align}
    \tr({\bf M}_{\mathrm{time}}^{k}{\bf B}^{-1}) &\leq \tr({\bf P}^{k}{\bf Q}^{H}({\bf B}^{k})^{-1}{\bf M}_{\mathrm{time}}^{k}({\bf B}^{k})^{-1}{\bf Q}{\bf P}^{k}{\bf P}^{-1}) \;.
    \label{eq:secondterm}
\end{align}
By defining $\mathbf{V}$ as
\begin{align}
    \mathbf{V} &=\left[\begin{array}{c}
    ({\bf B}^{k})^{-1}{\bf Q}{\bf P}^{k}{\bf P}^{\frac{-1}{2}}\\
    {\bf Q}{\bf P}^{\frac{1}{2}}
    \end{array}\right]\;, \label{eq:Udef}
\end{align}
the PSD property of $\mathbf{S}$ can be inferred as 
\begin{align}
    \mathbf{S} &= \left[\begin{array}{cc}
    ({\bf B}^{k})^{-1}{\bf Q}{\bf P}^{k}{\bf P}^{-1}{\bf P}^{k}{\bf Q}^{H}({\bf B}^{k})^{-1} & \textbf{I} \\
    \textbf{I} & {\bf Q}{\bf P}{\bf Q}^{H}
    \end{array}\right]=\mathbf{V}\mathbf{V}^{H}\geq0\;. \label{eq:Sdef}
\end{align}
Therefore by virtue of the Schur complement with ${\bf D}=({\bf B}^{k})^{-1}{\bf Q}{\bf P}^{k}{\bf P}^{-1}{\bf P}^{k}{\bf Q}^{H}({\bf B}^{k})^{-1}$, ${\bf G} = \textbf{I}$ and ${\bf B}={\bf Q}{\bf P}{\bf Q}^{H}$, we have:
\begin{equation}
     ({\bf B}^{k})^{-1}{\bf Q}{\bf P}^{k}{\bf P}^{-1}{\bf P}^{k}{\bf Q}^{H}({\bf B}^{k})^{-1} \geq \left({\bf Q}{\bf P}{\bf Q}^{H} \right)^{-1} \label{eq:afterSchur}\;.
\end{equation}
The inequality (Eq.~\eqref{eq:secondterm}) can be directly obtained by multiplying ${\bf M}_{\mathrm{time}}^{k}$ to both sides of Eq.~\eqref{eq:afterSchur}, applying the trace operator, using Eq.~\eqref{eq:FourierDiagonalization} and finally rearranging the terms within the trace operator:
\begin{equation}
     \tr({\bf M}_{\mathrm{time}}^{k}({\bf B}^{k})^{-1}{\bf Q}{\bf P}^{k}{\bf P}^{-1}{\bf P}^{k}{\bf Q}^{H}({\bf B}^{k})^{-1}) \geq \tr({\bf M}_{\mathrm{time}}^{k}{\bf B}^{-1}) \;. \label{eq:afterSchurfinal}
\end{equation}
Let ${\bf B}_{k}={\bf Q}{\bf P}^{k}{\bf Q}^{H}$ be the Fourier diagonalization of a fixed matrix ${\bf B}_{k}$ in the $k$-th iteration, one can derive an efficient update rule for the temporal covariance by rewriting Eq.~\eqref{eq:MinComstrainedTempo} and exploiting Propositions~\ref{prop:toeplitz} and Eq.~\eqref{eq:secondterm}: 
\begin{align}
    \tr(({\bf B}^{k})^{-1}{\bf B}) &+ \tr({\bf B}^{-1}{\bf M}_{\mathrm{time}}^{k}) \nonumber \\ &\leq  
    \tr(({\bf B}^{k})^{-1}{\bf Q}{\bf P}{\bf Q}^{H}) \nonumber + \tr({\bf P}^{k}{\bf Q}^{H}({\bf B}^{k})^{-1}{\bf M}_{\mathrm{time}}^{k}({\bf B}^{k})^{-1}{\bf Q}{\bf P}^{k}{\bf P}^{-1}) \nonumber \\ &= 
    \mathrm{diag}({\bf Q}^{H}({\bf B}^{k})^{-1}{\bf Q}){\bf p} + \mathrm{diag}({\bf P}^{k}{\bf Q}^{H}({\bf B}^{k})^{-1}{\bf M}_{\mathrm{time}}^{k}({\bf B}^{k})^{-1}{\bf Q}{\bf P}^{k}){\bf p}^{-1}\;, \label{eq:finalupdateTempII}
\end{align}
where ${\bf p}=\mathrm{vec}({\bf P})$, and ${\bf p}^{-1}$ is defined as the element-wise inversion of ${\bf p}$. 

We formulate the optimization problem as follows:
\begin{align}
    \mathcal{L}^{\mathrm{time}}_{\mathrm{toeplitz}}({\bf p}) &= \mathrm{diag}({\bf Q}^{H}({\bf B}^{k})^{-1}{\bf Q}){\bf p} \nonumber \\ +& \mathrm{diag}({\bf P}^{k}{\bf Q}^{H}({\bf B}^{k})^{-1}{\bf M}_{\mathrm{time}}^{k}({\bf B}^{k})^{-1}{\bf Q}{\bf P}^{k}){\bf p}^{-1}\;.
\label{eq:finalupdateTemp}
\end{align}
Let ${\bf W}^k:={\bf Q}^{H}({\bf B}^{k})^{-1}{\bf Q}$ and ${\bf O}^k:={\bf P}^{k}{\bf Q}^{H}({\bf B}^{k})^{-1}{\bf M}_{\mathrm{time}}^{k}({\bf B}^{k})^{-1}{\bf Q}{\bf P}^{k}$. Then, we rewrite $\mathcal{L}^{\mathrm{time}}_{\mathrm{toeplitz}}({\bf p})$ as   
\begin{align}
    \mathcal{L}^{\mathrm{time}}_{\mathrm{toeplitz}}({\bf p}) &= \mathrm{diag}({\bf W}^k){\bf p} + \mathrm{diag}({\bf O}^k){\bf p}^{-1}\;.
    \label{eq:time-diag-appendix}
\end{align}
The convexity of  $\mathcal{L}^{\mathrm{time}}_{\mathrm{toeplitz}}({\bf p})$ can be directly inferred from the convexity of $\mathrm{diag}\left[{\bf W}^k \right]{\bf p}$ and $\mathrm{diag}\left[{\bf O}^k\right]{\bf p}^{-1}$ with respect to ${\bf p}$ \citep[Chapter. 3]{boyd2004convex}.
\subsection{Part II: Derivation of the update rule in Eq.~\eqref{eq:BToeplitzUpdate}}
We now present the second part of the proof by deriving the update rule in Eq.~\eqref{eq:BToeplitzUpdate}.
Since the cost function $\mathcal{L}^{\mathrm{time}}_{\mathrm{toeplitz}}({\bf p})$ is convex, its optimal solution in the $k$-th iteration is unique. More concretely, a closed-form solution of the final update rule can be obtained by taking the derivative of Eq.~\eqref{eq:time-diag-appendix} with respect to ${\bf p}$ and setting it to zero:
\begin{align}
    p_{l}^{k+1} &\leftarrow\sqrt{\frac{\hat{g}_{l}^{k}}{\hat{z}_{l}^{k}}}\;\text{for}\;l=0,\hdots,L-1 \;, \text{where} \\ 
    \hat{\bf g} &=\mathrm{diag}({\bf P}^{k}{\bf Q}^{H}({\bf B}^{k})^{-1}{\bf M}_{\mathrm{time}}^{k}({\bf B}^{k})^{-1}{\bf Q}{\bf P}^{k}) \\
    \hat{\bf z} &=\mathrm{diag}({\bf Q}^{H}({\bf B}^{k})^{-1}{\bf Q})\;,
\end{align}
which concludes the proof.
\end{proof}

\section{Proof of Theorem~\ref{theo:eff-post}}
\label{appendix:eff-post}
\begin{proof}

The proof is inspired by ideas presented in \citet{saatcci2012scalable,rakitsch2013all,wu2018learning} for spatio-temporal Gaussian process inference, and parallels the one proposed in \citet{solin2016regularizing}. \citet{rakitsch2013all,wu2018learning} provide an efficient method for computing the non-convex spatio-temporal ML cost function by exploiting the compatibility between diagonalization and the Kronecker product. Here we use similar ideas to obtain the posterior mean in an efficient way.

Recalling the diagonalization of the temporal correlation matrix as ${\bf B}={\bf Q}{\bf P}{\bf Q}^{H}$ and considering the eigenvalue decomposition of ${\bf L}\bm{\Gamma}{\bf L}^{\top}$ as ${\bf L}\bm{\Gamma}{\bf L}^{\top}= {\bf U}_{\bf x}{\bf D}_{\bf x}{\bf U}_{\bf x}^{\top}$ with ${\bf D}_{\bf x}=\mathrm{diag}(d_1,\hdots,d_M)$, we have: 
\begin{align}
    \bar{\bf x}_{g} & =\left(\bm{\Gamma}\otimes{\bf B}\right){\bf D}^{\top}\Tilde{\bm \Sigma}_{\bf y}^{-1}{\bf y}_{g} \nonumber \\
     & =\left(\bm{\Gamma}\otimes{\bf B}\right)({\bf L}\otimes{\bf I})^{\top}\left({\bm \Lambda} \otimes {\bf B}+{\bf D}{\bm \Sigma}_{0}{\bf D}^{\top}\right)^{-1}\mathrm{vec}({\bf Y}_{g}^{\top}) \nonumber \\
     & =\left(\bm{\Gamma}\otimes{\bf B}\right)({\bf L}^{\top}\otimes{\bf I})\left(({\bm \Lambda}+{\bf L}\bm{\Gamma}{\bf L}^{\top})\otimes{\bf B}\right)^{-1}\mathrm{vec}({\bf Y}_{g}^{\top}) \nonumber \\
     & =\left(\bm{\Gamma}{\bf L}^{\top}\otimes{\bf B}\right)\left(({\bm \Lambda}+{\bf L}\bm{\Gamma}{\bf L}^{\top})\otimes{\bf B}\right)^{-1}\mathrm{vec}({\bf Y}_{g}^{\top}) \nonumber \\
     & =\left(\bm{\Gamma}{\bf L}^{\top}{\bf U}_{{\bf x}}\otimes{\bf Q}{\bf P}\right)({\bm \Omega})^{-1}({\bf U}_{{\bf x}}^{\top} \otimes {\bf Q}^{H})\mathrm{vec}({\bf Y}_{g}^{\top}) \nonumber \\
     & =\left(\bm{\Gamma}{\bf L}^{\top}{\bf U}_{{\bf x}}\otimes{\bf Q}{\bf P}\right)({\bm \Omega})^{-1}\tr\left({\bf Q}^{H}{\bf Y}_{g}^{\top}{\bf U}_{{\bf x}}\right) \nonumber \\
     & =\left(\bm{\Gamma}{\bf L}^{\top}{\bf U}_{{\bf x}}\otimes{\bf Q}{\bf P}\right)\tr\left({\bm \Pi} \odot{\bf Q}^{H}{\bf Y}_{g}^{\top}{\bf U}_{\bf x}\right) \nonumber \\
     & = \tr\left({\bf Q}{\bf P} \left({\bm \Pi} \odot{\bf Q}^{H}{\bf Y}_{g}^{\top}{\bf U}_{\bf x}\right)\left({\bf U}_{{\bf x}}^{\top}{\bf L}\bm{\Gamma}^{\top}\right) \right) \;,
\end{align}
where $\odot$ denotes the Hadamard product between corresponding elements of two matrices. ${\bm \Omega}$ and ${\bm \Pi}$ are defined as follows: ${\bm \Omega}= {\bm \Lambda} + {\bf D}_{\bf x} \otimes {\bf P}$ and $[{\bm \Pi}]_{l,m}= \frac{1}{{\sigma}_m^2 + p_l d_m}\;\text{for}\; l=1,\hdots,L;\;m=1,\hdots,M$. Note that the last four lines are derived based on the following matrix equality: 
\begin{align}
    \tr({\bf A}^{\top}{\bf B}{\bf C}{\bf D}^{\top})=\mathrm{vec}({\bf A})^{\top}({\bf D}\otimes {\bf B})\mathrm{vec}({\bf C}).
\end{align}
Together with the update rule in Eq.~\eqref{eq:efficientPosterior}, this concludes the proof of Theorem~\ref{theo:eff-post}.
\end{proof}
\section{Details on the simulation set-up}
\label{appendix-Simulation}
\subsection{Forward modeling}
Populations of pyramidal neurons in the cortical gray matter are known to be the main drivers of the EEG signal \citep{hamalainen1993magnetoencephalography,baillet2001electromagnetic}. Here, we use a realistic volume conductor model of the human head to model the linear relationship between primary electrical source currents generated within these populations and the resulting scalp surface potentials captured by EEG electrodes. The lead field matrix, ${\bf L} \in \mathbb{R}^{58 \times 2004}$, which serves as the forward model in our simulations, was generated using the New York Head model \citep{huang2016new}. The New York Head model provides a segmentation of an average human head into six different tissue types, taking into account the realistic anatomy and electrical
tissue conductivities. In this model, 2004 dipolar current sources were placed evenly on the cortical surface and 58 sensors were placed on the scalp according to the extended 10-20 system \citep{oostenveld_five_2001}. Finally, the lead field matrix was computed using the finite element method (FEM) for a given head geometry and exploiting the quasi-static approximation of Maxwell's equations
\citep{hamalainen1993magnetoencephalography,baillet2001electromagnetic,gramfort2009mapping,huang2016new}.

Note that in accordance with the predominant orientation of pyramidal neuron assemblies, the orientation of all simulated source currents was fixed to be perpendicular to the cortical surface, so that only the scalar deflection of each source along the fixed orientation needs to be estimated. In real data analyses in Section~\ref{sec:real-data-analysis} and Appendix~\ref{appendix:real-data}, however, surface normals are hard to estimate or even undefined in case of volumetric reconstructions. Consequently, we model each source in real data analyses as a full 3-dimensional current vector. This is achieved by introducing three variance parameters for each source within the source covariance matrix, ${\bm \Gamma}^{3\text{D}} =\mathrm{diag}({\bm \gamma}^{3\text{D}}) = [\gamma^{x}_{1},\gamma^{y}_{1},\gamma^{z}_{1},\dots,\gamma^{x}_{N},\gamma^{y}_{N},\gamma^{z}_{N}]^\top$. As all algorithms considered here model the source covariance matrix ${\bm \Gamma}$ to be diagonal, this extension can be readily implemented. Correspondingly, a full 3D leadfield matrix, ${\bf L}^{3\text{D}} \in \mathbb{R}^{M\times 3N}$, is used.

\subsection{Pseudo-EEG signal generation}
We simulated a sparse set of $N_0 =3$ active sources, which were placed at random positions on the cortex. To simulate the electrical neural activity of these sources, $T=\{10,20,50,100\}$
time points were sampled from a univariate linear autoregressive (AR) process, which models the activity at time $t$ as a linear combination of the $P$ past values:
\begin{align}
    {x}_{i}(t) = \sum_{p=1}^{P}a_{i}(p){x}_{i}(t-p) + \xi_{i}(t),\; \mathrm{for}\;i=1,2,3 \;.
\end{align}
Here, $a_{i}(p)$ for $i=1,2,3$ are linear AR coefficients, and $P$ is the order of the AR model. The model residuals $\xi_{i}(\cdot)$ for $i=1,2,3$ are also referred to as the innovation process; their variance determines the stability of the overall AR process. We here assume uncorrelated standard normal distributed innovations, which are independent for all sources. In this experiment, we use stable AR systems of order $P=\{1,2,5,7\}$. The resulting source distribution, represented as $\mathbf{X} = [\mathbf{x}(1),\hdots,\mathbf{x}(T)]$, was projected to the EEG sensors through application of lead field matrix: ${\bf Y}^{\mathrm{signal}}= \mathbf{L}\mathbf{X}$. Next we added Gaussian white noise to the sensor space signal. To this end, the same number of data points as the sources were sampled from a zero-mean normal distribution, where the time points assumed to be independent and identically distributed. The resulting noise distribution, represented as ${\bf E}=[{\bf e}(1),\dots,{\bf e}(T)]$, is then normalized by its Frobenius norm and added to the signal matrix ${\bf Y}^{\mathrm{signal}}$ as follows: ${\bf Y}={\bf Y}^{\mathrm{signal}}+\frac{(1-\alpha)\left\Vert {{\bf Y}^{\mathrm{signal}}}\right\Vert_{F}}{\alpha \left\Vert{\bf E}\right\Vert_{F}}{\bf E}$, where $\alpha$ determines the signal-to-noise ratio (SNR) in sensor space. Precisely, SNR is defined as follows: $\mathrm{SNR} = 20\mathrm{log}_{10}\left(\nicefrac{\alpha}{1-\alpha}\right)$. In this experiment the following values of $\alpha$ were used: $\alpha$=\{0.55, 0.65, 0.7, 0.8\}, which correspond to the following SNRs: SNR=\{1.7, 5.4, 7.4, 12\}~(dB). Interested reader can refer to \citet{haufe2016simulation} for a more details on this simulation framework.

\subsection{Source reconstruction and evaluation metrics}
We applied thin Dugh to the synthetic datasets described above. In addition to thin Dugh, one further Type-II Bayesian learning scheme, namely Champagne \citep{wipf2010robust}, and two Type-I source reconstruction schemes, namely S-FLEX \citep{haufe2011large} and eLORETA \citep{pascual2007discrete}, were also included as benchmarks with respect to source reconstruction performance. S-FLEX is used as an example of a sparse Type-I Bayesian learning method based on $\ell_1$-norm minimization. As spatial basis functions, unit impulses were used, so that the resulting estimate was identical to the so-called minimum-current estimate (MCE) \citep{matsuura1995selective}. eLORETA estimate, as an example of a smooth inverse solution based on weighted $\ell_2^2$-norm minimization, was used with $5\%$ regularization, whereas S-FLEX was fitted so that the residual variance was consistent with the ground-truth noise level. Note that the $5\%$ rule is chosen as it gives the best performance across a subset of regularization values ranging between $0.5\%$ to $15\%$. For thin Dugh, the noise variances as well as the variances of all voxels were initialized randomly by sampling from a standard normal distribution. The optimization program was terminated after reaching convergence. Convergence was defined if the relative change of the Frobenius-norm of the reconstructed sources between subsequent iterations was less than $10^{-8}$. A maximum of 1000 iterations was carried out if no convergence was reached beforehand.

Source reconstruction performance was evaluated according to two different measures, the \emph{earth mover's distance} (EMD), used to quantify the spatial localization accuracy, and the correlation between the original and reconstructed sources, $\hat{{\bf X}}$ and ${\bf X}$. The EMD metric measures the cost needed to map two probability distributions, defined on the same metric domain, into each other, see \citep{rubner2000emd,haufe2008combining}. It was applied here to the power of the true and estimated source activations defined on the cortical surface of the brain, which were obtained by taking the voxel-wise $\ell_2$-norm along the time domain. EMD was normalized to $[0, 1]$. The correlation between simulated and reconstructed source time courses was assessed as the mean of the absolute correlations obtained for each source, after optimally matching simulated and reconstructed sources. To this end, Pearson correlation between all pairs of simulated and reconstructed (i.e., those with non-zero activations) sources was measured. Each simulated source was matched to a reconstructed source based on maximum absolute correlation. Time-course correlation error (TCE) was then defined as one minus the average of these absolute correlations across sources. Each simulation was carried out 100 times using different instances of $\bf{X}$ and $\bf E$, and the mean and standard error of the mean (SEM) of each performance measure across repetitions was calculated.
\section{Real data analysis}
\label{appendix:real-data}
\subsection{Auditory and visual evoked fields (AEF and VEF)}
The MEG data used in this article were acquired in the Biomagnetic Imaging Laboratory at the University of California San Francisco (UCSF) with a CTF Omega 2000 whole-head MEG system from VSM MedTech (Coquitlam, BC, Canada) with $1200$ Hz sampling rate. The neural responses for one subject's auditory evoked fields (AEF) and visual evoked fields (VEF) were localized. The AEF response was elicited while subjects were passively listening to $600$ ms duration tones ($1$ kHz) presented binaurally. Data from 120 trial epochs were analysed. The VEF response was elicited while subjects were viewing images of objects projected onto a screen and subjects were instructed to name the objects verbally. Both AEF and VEF data were first digitally filtered from $1$ to $70$ Hz to remove artifacts and DC offset, time-aligned to the stimulus. Different number of trials were included for algorithm analyses. The pre-stimulus window was selected to be $-100$ ms to $-5$ ms and the post-stimulus time window was selected to be $60$ ms to $180$ ms, where $0$ ms is the onset of the tone. The lead field for each subject was calculated with NUTMEG \citep{Dalal:2004} using a single-sphere head model (two spherical orientation lead fields) and an 8~mm voxel grid. Each column was normalized to have a norm of unity. Further details on these datasets can be found in \citep{wipf2010robust,dalal2011meg,owen2012performance,cai2019robust}.

\begin{figure}
  \centering
  \centerline{\includegraphics[width=\textwidth,trim=0 15cm 0 0,clip]{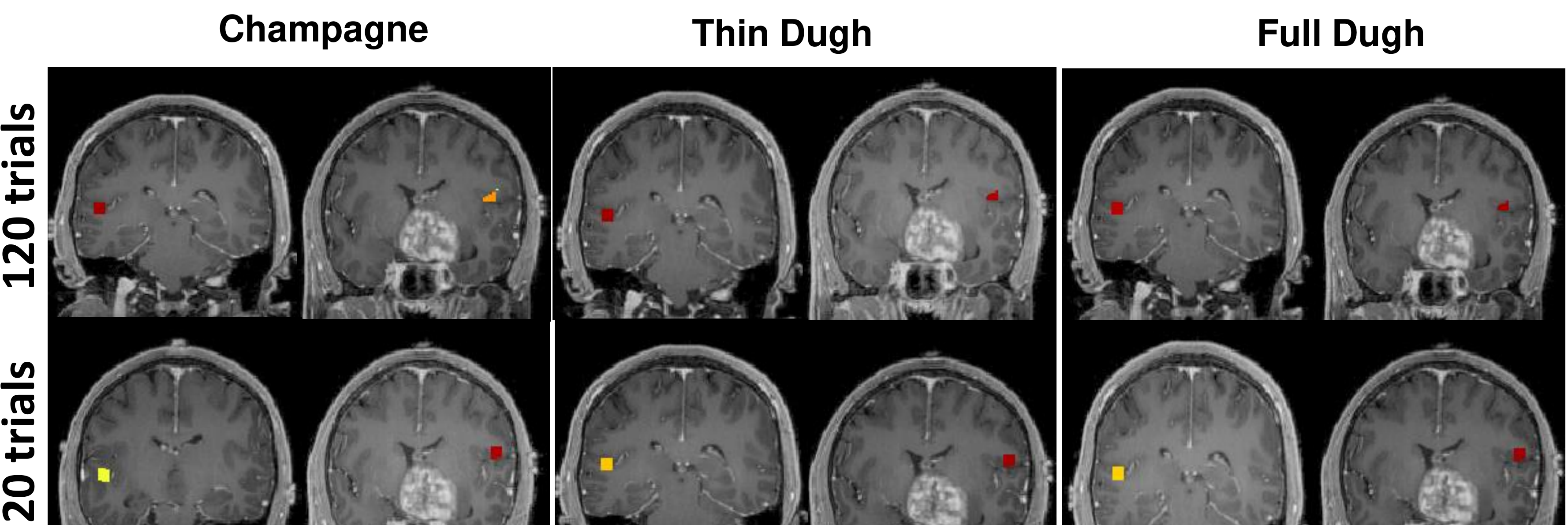}}
\caption{Robustness of Dugh and Champagne performance when the number of trials is increased to 20 and 120.}
\label{fig:AEF-appendix-Trials}
\vspace{-2mm}
\end{figure}

\begin{figure}
 \centering
 \centerline{\includegraphics[width=\textwidth]{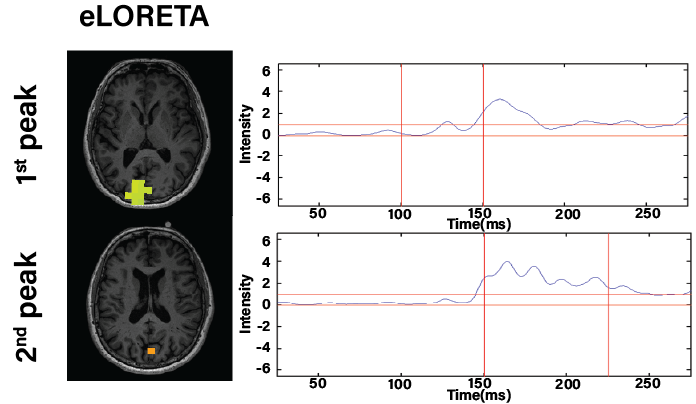}}
\caption{VEF performance of benchmark algorithm eLORETA. This benchmark did not yield reliable results for 5 trial epochs. Even when the number of trials were increased to 20, benchmark's performance yielded neither good spatial localization of the two visual cortical areas nor good estimation of the time courses of these activations.}
\label{fig:VEF-appendix}
\vspace{-2mm}
\end{figure}

\ali{Figure~\ref{fig:AEF-appendix-Trials} shows the reconstructed sources of the Auditory Evoked Fields (AEF) from a representative subject using Champagne, thin and full Dugh. In this case, we tested the reconstruction performance of all algorithms with the number of trials limited to 20 and 120. As Figure~\ref{fig:AEF-appendix-Trials} demonstrates, the performance of Dugh remains robust as the number of trials is increased to 20 and 120 in Figure~\ref{fig:AEF-appendix-Trials}. Finally, the VEF performance of benchmark algorithm eLORETA is demonstrated in Figure~\ref{fig:VEF-appendix}.
}
%
%


\subsection{EEG Data: Faces vs scrambled pictures}
\begin{figure}[h]
  \centering
  \centerline{\includegraphics[width=\textwidth]{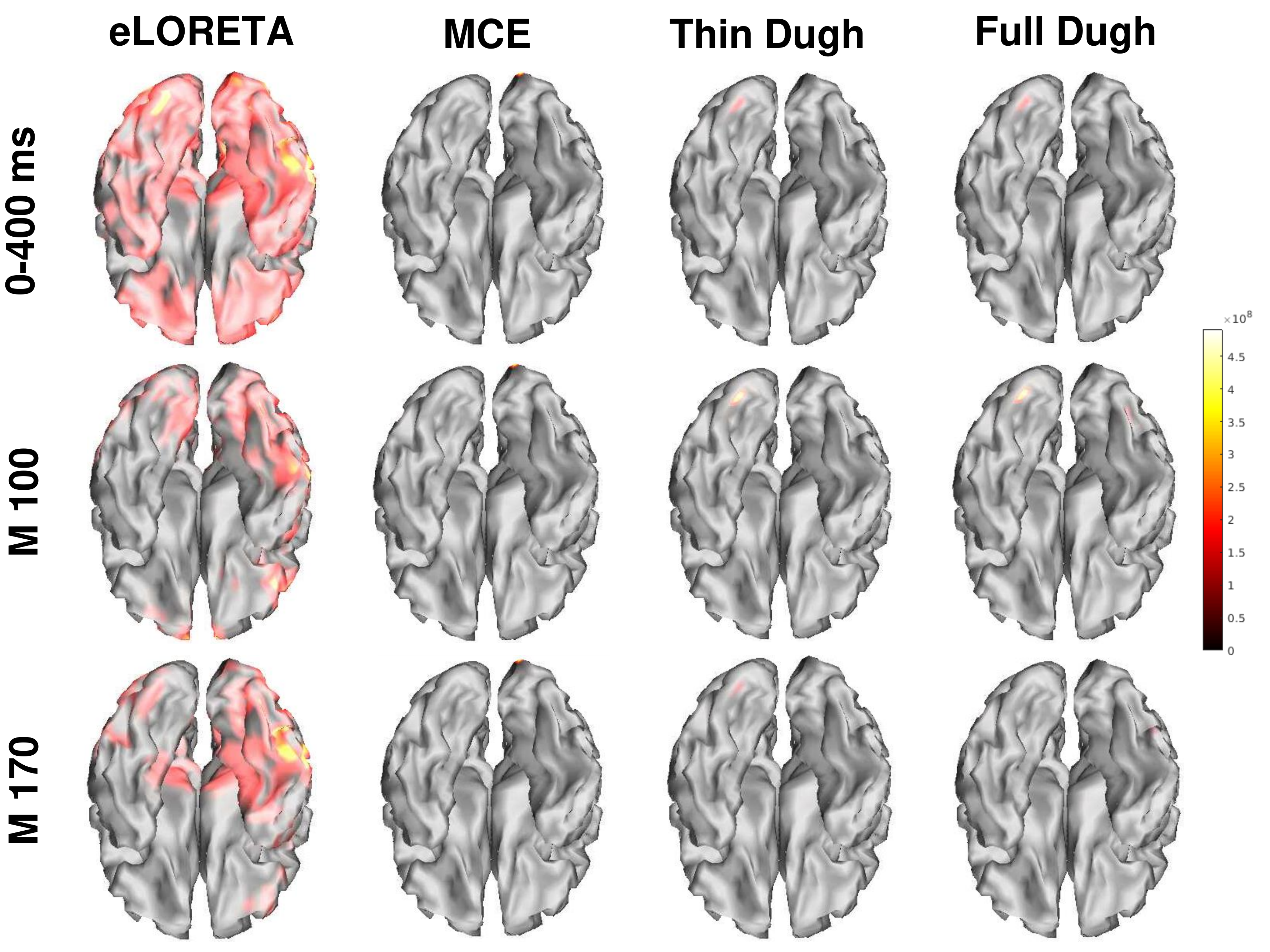}}
\caption{Performance of Dugh and benchmarks on EEG data acquired during a face recognition task. Dugh was able to provide more focal and distinct activations for the M100 and M170 responses that were not clearly identified using the benchmarks eLORETA and MCE.}
\label{fig:EEG}
\end{figure}

A publicly available EEG dataset (128-channel Biosemi ActiveTwo system) was downloaded from the SPM website (http://www.fil.ion.ucl.ac.uk/spm/data/mmfaces) and the
lead field was calculated in SPM8 using a three-shell
spherical model at the coarse resolution of 5124 voxels
at approximately 8 mm spacing. These EEG data were also obtained during a visual response paradigm that involved randomized presentation of at least 86 faces and 86 scrambled faces. To examine the differential responses to faces across all trials, the averaged responses to scrambled-faces were
subtracted from the averaged responses to faces. The result is demonstrated in Figure~\ref{fig:EEG}.

\end{document}